\documentclass[pdflatex,sn-mathphys-num]{sn-jnl}
\usepackage{amsmath,amsfonts,amssymb,amsthm}
\usepackage{bbm}
\usepackage{stackengine}
\usepackage{algorithm}
\usepackage{algpseudocode}
\algrenewcommand\textproc{}
\usepackage{listings}
\usepackage{epsfig}
\usepackage{color} 
\usepackage{graphicx}
\usepackage{subfigure}
\usepackage{float}
\usepackage{hyperref}
\usepackage{footnote}
\usepackage{mathrsfs}
\usepackage{booktabs}
\usepackage{caption}
\usepackage[section]{placeins}
\usepackage{diagbox}
\usepackage{tabularx}
\usepackage{multirow}
\usepackage{makecell}
\usepackage{array}
\usepackage{appendix}

\theoremstyle{thmstyleone}%
\theoremstyle{thmstyletwo}%
\theoremstyle{thmstylethree}%

\newcommand{\bx}{\mathbf x}
\newcommand{\by}{\mathbf y}
\newcommand{\be}{\mathbf e}
\newcommand{\bz}{\mathbf z}
\newcommand{\br}{\mathbf r}
\newcommand{\bbR}{\mathbb R}

\begin{document}
\title{A level-wise training scheme for learning neural multigrid smoothers with application to integral equations}

\author*[1]{\fnm{Lingfeng} \sur{Li}}\email{lilingfeng@himis-sz.cn}
\author[2]{\fnm{Yin King} \sur{Chu}}
\author[3]{\fnm{Raymond} \sur{Chan}}
\author[4]{\fnm{Justin} \sur{Wan}}

\affil[1]{\orgname{Hetao Institute of Mathematics and Interdisciplinary Sciences, \orgaddress{\city{Shenzhen}, \state{Guangdong}, \country{China}}}}

\affil[2]{\orgdiv{Department of Mathematics}, \orgname{Hong Kong University of Science and Technology}, \orgaddress{\city{Clear Water Bay}, \state{Hong Kong}, \country{China}}}

\affil[3]{\orgdiv{Department of Data Science, Department of Operations and Risk Management},\orgname{Lingnan University}, \orgaddress{\city{Tuen Mun}, \state{Hong Kong}, \country{China}}}

\affil[4]{\orgdiv{David R. Cheriton School of Computer Science}, \orgname{University of Waterloo}, \orgaddress{\city{Waterloo},  \state{Ontario}, \country{Canada}}}

\abstract{
    Convolution-type integral equations commonly occur in signal processing and image processing. Discretizing these equations yields large and ill-conditioned linear systems. While the classic multigrid method is effective for solving linear systems derived from partial differential equations (PDE) problems, it fails to solve integral equations because its smoothers, which are implemented as conventional relaxation methods, are ineffective in reducing high-frequency components in the errors.
    We propose a novel neural multigrid scheme where learned neural operators replace classical smoothers. Unlike classical smoothers, these operators are trained offline. Once trained, the neural smoothers generalize to new right-hand-side vectors without retraining, making it an efficient solver. We design level-wise loss functions incorporating spectral filtering to emulate the multigrid frequency decomposition principle, ensuring each operator focuses on solving distinct high-frequency spectral bands.
    Although we focus on integral equations, the framework is generalizable to all kinds of problems, including PDE problems. Our experiments demonstrate superior efficiency over classical solvers and robust convergence across varying problem sizes and regularization weights.
}


\maketitle

\section{Introduction}
The multigrid method \cite{xu2017algebraic} is a highly efficient numerical solver for linear systems arising from partial differential equations (PDEs). It typically consists of two components: 
% the smoothing step and the coarse-grid correction step.
The smoothing step, performed on fine grids by classical relaxation methods (e.g., Jacobi), dampens geometric high-frequency components (oscillatory parts) of the errors. The coarse-grid correction, implemented on coarse grids via direct solvers (e.g., Gaussian elimination), dampens low-frequency components (smooth parts) of the errors. Since the direct solver operates on a significantly coarser grid than the original, the computational cost is manageable. 

However, this method often fails for certain linear systems arising from integral equations, especially convolution-type integral equations with smooth kernels \cite{chan1997multigrid}.
The reason is that conventional relaxation methods cannot solve high-frequency components for these equations efficiently, making the smoothing steps fail. 
Developing effective smoothing steps for these types of integral equations to address high-frequency components remains a challenging task.

In this work, we train neural operators to target high-frequency error components in integral equations. Neural operator solvers are neural networks designed to learn solution mappings for mathematical problems. Though the training time of neural operators is expensive, once it is trained, it can efficiently make predictions according to different input conditions.
They have been integrated into multigrid schemes for solving linear systems arising from PDEs in recent years \cite{hu2024hybrid,kahana2023geometry,zhang2022hybrid, Zhang2024BlendingNO,huang2022learning,song2025physics}. One line of research \cite{hu2024hybrid,kahana2023geometry,zhang2022hybrid, Zhang2024BlendingNO} replaces the coarse-grid correction step with neural operators. These approaches combine neural networks with classical relaxation methods to form hybrid iterative solvers. They perform smoothing steps (by relaxation methods) on the finest grid only and employ neural operators for corrections on the same grid, without interpolation or restriction operations between different grid levels. Another line of research \cite{huang2022learning,song2025physics} substitutes smoothers on each grid level with neural operators while retaining the multilevel structure characteristic of conventional multigrid methods. Nevertheless, all these studies focus exclusively on solving PDE problems, which classical multigrid methods can also solve efficiently.

A recent work \cite{li2025solving} adopts a similar framework as \cite{Zhang2024BlendingNO}, \textit{i.e.}, replacing the coarse-grid correction step in the multigrid scheme, to address integral equations. They train neural operators to focus on reducing high-frequency components in the errors by minimizing the residuals of preconditioned systems. By combining these trained neural operators with relaxation methods, they construct an efficient hybrid iterative solver that addresses both high- and low-frequency components. This solver demonstrates effectiveness in addressing ill-conditioned linear systems derived from one- and two-dimensional integral equations, outperforming classical methods such as conventional multigrid and preconditioned conjugate gradient.

Unlike \cite{li2025solving}, we propose to replace relaxation methods across multigrid levels with neural operators, which align with methodologies in \cite{huang2022learning,song2025physics}.
For each grid level, we independently construct and train a neural operator to perform the smoothing step. A novel level-wise training loss function, which involves specific high-frequency filters, is designed to ensure the neural operators solve high-frequency components on their corresponding grids and leave low-frequency components to coarser grid levels. In \cite{li2025solving}, training neural operators needs a preconditioning matrix with spectral properties inverse to those of the coefficient matrix, and constructing such a preconditioning matrix requires certain prior knowledge about the coefficient matrix. In contrast, our proposed method does not need any preconditioning matrix, making it more generalizable to different linear systems. Experimental results show that our neural multigrid solver exhibits high efficiency, robustness, and flexibility in solving integral equations. Its convergence rate is robust to different problem sizes and regularization weights, and the coarsest grid level can be adjusted after training without compromising the convergence rate.

\section{Preliminaries}
\subsection{Convolution-type integral equations}
In image processing and data analysis applications \cite{chui1982application,grenander1958toeplitz,king1989digital}, we often need to solve integral equations of the form:
\begin{equation}
    \alpha \text{R}(u)(z)+\int_\Omega \mathcal{K}(z-z')u(z')dz'=f(z),\quad z\in[0,1]. \label{eq:integral equation}
\end{equation}
where $\mathcal{K}$ is a smooth convolution kernel, such as a Gaussian smoothing kernel, and $\alpha$ is the regularization weight parameter. The regularization term $R(u)$ is employed to handle the ill-condition \cite{beck2009fast}. Consider a uniform mesh $\Omega_h$ partitioning $[0,1]$ into $n$ equally spaced cells with mesh size $h = 1/n$. Under appropriate boundary conditions, discretizing \eqref{eq:integral equation} on $\Omega_h$ yields a linear system  
$$ A\mathbf{x} = \mathbf{y}, $$  
where $A$ is the coefficient matrix that has some special structures like Toeplitz or Block Toeplitz with Toeplitz Block (BTTB). A matrix is called Toeplitz if each diagonal is a constant, and a matrix is called BTTB if it has the form
\begin{equation*}
    \left[\begin{array}{cccccc}
A_0 & A_{1} & A_{2} & & \dots & A_n  \\
A_{-1} & A_0 & A_{1} & \ddots & &  \\
A_{-2} & A_{-1} & \ddots & \ddots &  &  \\
& & \ddots& \ddots & A_1 & A_2\\
 &  & \ddots & A_{-1} & A_{0} & A_{1} \\
 A_{-n} & \dots  &  & A_{-2} & A_{-1} & A_{0} 
\end{array}\right],
\end{equation*}
where each $A_i$, $i=0,\pm1,\pm2,\dots$, is a Toeplitz matrix.
The solution of a Toeplitz system can be found efficiently by leveraging its special structure. Many fast solvers, including direct methods \cite{ammar1988superfast,bitmead1980asymptotically,brent1980fast} and iterative methods \cite{chan1994circulant,chan1988optimal,tyrtyshnikov1992optimal,chan1992circulant}, have been proposed for solving Toeplitz systems. 

\subsection{Classical relaxation methods} \label{sec:smoothers}
Classical relaxation methods, such as the Jacobi method and the Gauss-Seidel method, offer a straightforward and systematic approach to solving linear systems of the form \( A \bx = \by \) by iteratively refining an approximate solution. For example, the Jacobi method decomposes the matrix \( A \) into two parts \( D\) and \(A-D\) where $D$ is the diagonal part of $A$. It iteratively updates the solution by \(\bx^{(\nu+1)}=\bx^{(\nu)}+D^{-1}(\by-A\bx^{(\nu)})\) where \(\nu=0,1,\dots\) and $\bx^{(0)}$ is an initial guess. The generated sequence $\{\bx^{(\nu)}\}$ can converge to the true solution under appropriate conditions. However, their convergence rate would depend on the condition number of $A$. 

For most PDE problems, these relaxation methods possess an error-smoothing effect. They solve the high-frequency components, \textit{i.e.}, vectors with oscillatory entries, much more efficiently than low-frequency components \cite{xu2017algebraic}, \textit{i.e.}, vectors with smooth entries, making the remaining error less oscillatory and smoother. 

\begin{figure}
    \centering
    \includegraphics[width=0.4\linewidth]{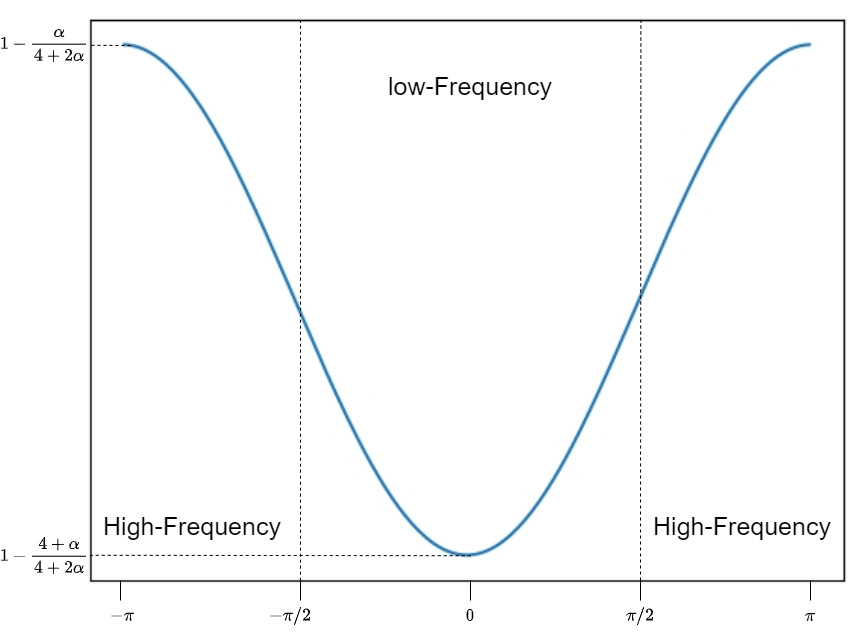}
    \caption{Plot of $1-\mu_\phi$ for $\phi\in[-\pi,\pi]$ when $\omega=1/2$.}
    \label{fig:smoothing factor}
\end{figure}

To further explain the smoothing effect, we can consider solving \( A \bx = \by \) with the following coefficient matrix:
\[
A = \begin{bmatrix}
2+\alpha & -1 & & & -1\\
-1 & 2+\alpha & -1 & &\\
& \ddots & \ddots & \ddots\\
& & -1 & 2+\alpha & -1 \\
-1 & & & -1 & 2+\alpha
\end{bmatrix}\in\bbR^{n\times n},
\]
Here, $A$ is discretized from a simple second-order PDE with periodic boundary conditions:
\[-u''(z)+\alpha u(z)=f(z),\quad z\in[0,1].\]
% The eigenvectors of $A$ are $\lambda_k(A)=4\sin^2\left(\frac{k\pi}{4n}\right),\quad k=1,\dots,n,$
% and the eigenvector corresponding to $\lambda_k(A)$ is
% \[ \mathbf v_k=\left[\sin\left(\frac{1k\pi}{n}\right),\sin\left(\frac{2k\pi}{n}\right),\dots,\sin\left(\frac{nk\pi}{n}\right)\right]. \]
Any vector in $\bbR^n$ can be decomposed using a set of Fourier basis $\{\be_{\phi_k}\}_{k=0}^{n-1}$ where
\[ \be_{\phi_k}=\left[\exp\left(-i\phi_k0\right),\exp\left(-i\phi_k1\right),\dots,\exp\left(-i\phi_k(n-1)\right)\right],\quad \phi_k=\frac{(2k-n)\pi}{n} \]
Here, $\phi_k\in[-\pi,\pi]$ for all $k=0,\dots,n-1$, so we simply view $\phi_k$ as a continuous variable in $[-\pi,\pi]$ and denote it as $\phi$. Typically, $\be_{\phi}$ is called the low-frequency mode if $\phi\in [-\frac{\pi}{2},\frac{\pi}{2}]$, and it is called the high-frequency mode otherwise \cite{briggs2000multigrid}.
We consider the weighted Jacobi method for solving $A\bx=\by$:
\[
\bx^{(\nu+1)} = \bx^{(\nu)} + \omega D^{-1}(\by - A \bx^{(\nu)}), \quad D = \text{diag}(A).
\]
The error at the $(\nu+1)$-th step is $\bx^{(\nu+1)}-\bx$ and it can be written as
\[ \bx^{(\nu+1)}-\bx=S_n(\bx^{(\nu)}-\bx), \]
where \( S_n = I - \frac{\omega}{2+\alpha} A \) is the error propagation matrix, and $I$ is an identity matrix. 
Multiplying \( A\) with a Fourier mode \( \be_{\phi} \) yields
\[
A \be_\phi = \left((2+\alpha) - 2\cos(\phi+\pi)\right) \be_\phi,
\]
and consequently \(S_n\be_\phi=\mu_\phi\be_\phi,\)
where $\mu_\phi:=1-\frac{\omega}{2+\alpha}(2+\alpha - 2\cos(\phi+\pi))$. This demonstrates that one iteration of the weighted Jacobi method reduces the component $\mathbf{e}_\phi$ in the error $\mathbf{x}^\nu - \mathbf{x}$ to $\mu_\phi$ of its original magnitude, achieving a reduction factor of $1 - \mu_\phi$. By selecting an appropriate weight parameter $\omega$, we ensure $\mu_\phi \in (0,1)$ for all $\phi \in [-\pi, \pi]$. 
We visualize the reduction factor $1-\mu_\phi$ for $\phi\in[-\pi,\pi]$ when $\omega=1/2$ in Figure \ref{fig:smoothing factor}. We see that high-frequency modes dampen much faster than low-frequency modes. The reduction of the lowest frequency mode ($\phi=0$) is $1-\frac{4+\alpha}{4+2\alpha}$, which is close to $0$ when $\alpha$ is small, while the reduction of the highest frequency mode ($\phi=\pm\pi$) is $1-\frac{\alpha}{4+2\alpha}$, which is close to $1$ when $\alpha$ is small.

\begin{figure}
    \centering
    \includegraphics[width=0.8\textwidth]{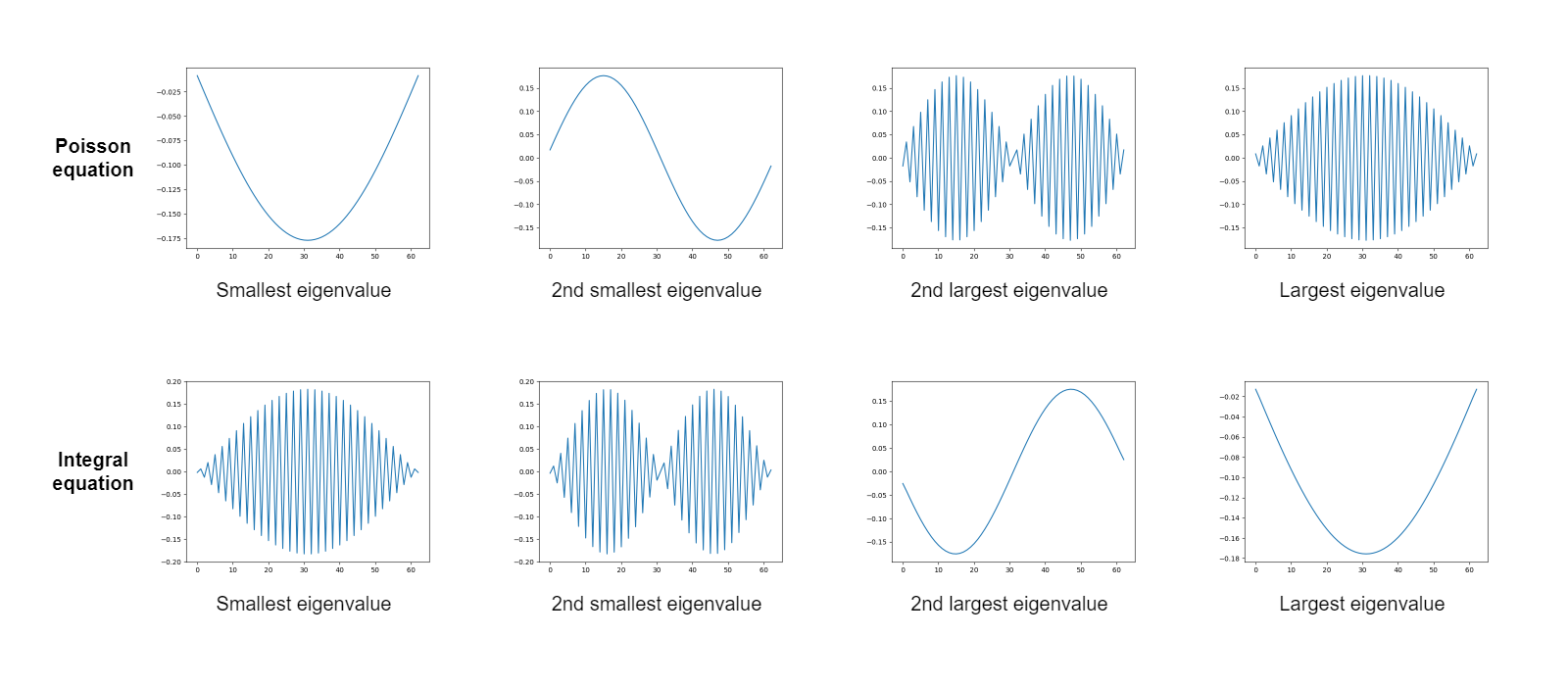}
    \caption{Eigenvectors corresponding to different eigenvalues of the linear system discretized from the Poisson equation (first row) and the Integral equation (second row).}
    \label{fig:eigenvectors}
\end{figure}

\begin{figure}
        \centering
        \subfigure[Poisson equation]{\includegraphics[width=0.4\linewidth]{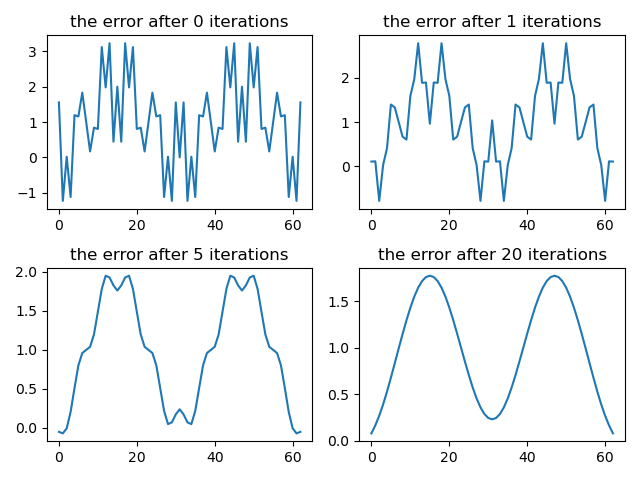}}
        \subfigure[Integral equation]{\includegraphics[width=0.4\linewidth]{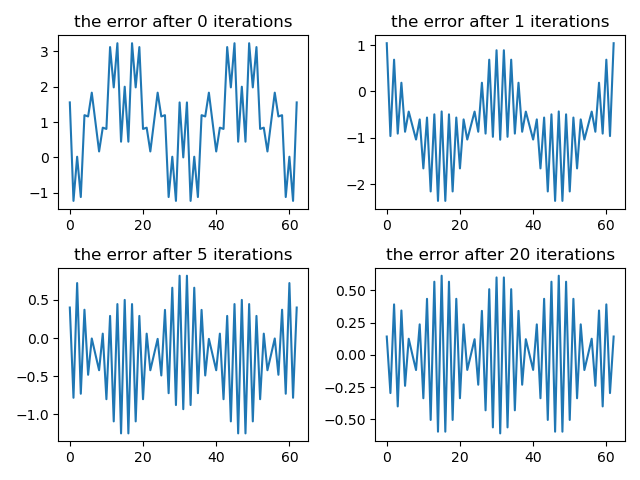}}
        \caption{Error vectors when solving different linear systems using the Jacobi method.}
        \label{fig:jacobi_iterations}
    \end{figure}

In general, relaxation methods can solve error components that correspond to eigenvectors of large eigenvalues of the coefficient matrix $A$ more effectively compared to components that correspond to eigenvectors of small eigenvalues of $A$. For most PDE problems, eigenvectors of large eigenvalues are high-frequency, \textit{i.e.}, oscillatory, while eigenvectors of small eigenvalues are low-frequency, \textit{i.e.}, smooth. Consequently, relaxation methods can solve high-frequency components and smooth the error very fast.
However, for convolution-type integral equations, relaxation methods can not effectively smooth errors, because their coefficient matrices exhibit inverse spectral properties, \textit{i.e.}, eigenvectors of large eigenvalues are smooth while eigenvectors of small eigenvalues are oscillatory. The spectral difference of two types of problems is shown in Figure \ref{fig:eigenvectors}, where we compare the eigenvectors corresponding to the largest and smallest eigenvalues of the coefficient matrices for a Poisson equation and an integral equation (\ref{eq:integral equation}) with $\mathcal K$ defined as a Gaussian smoothing kernel. We further examine the evolution of error vectors while solving the Poisson equation and the integral equation using the Jacobi method in Figure \ref{fig:jacobi_iterations}. The Jacobi method always solves the error components corresponding to eigenvectors of large eigenvalues of the underlying coefficient matrix efficiently: it damps geometric high-frequency components efficiently for the Poisson equation, while damps geometric low-frequency components efficiently for the Integral equation. 

\subsection{Multigrid method}
The multigrid (MG) method is a hierarchical numerical scheme. While relaxation methods like Jacobi are efficient at removing oscillatory (high-frequency) error, they require thousands of iterations to eliminate smooth (low-frequency) error. The MG method addresses this by viewing the problem at multiple scales. It iteratively refines the solution using multigrid cycles and reduces both high-frequency and low-frequency components in the errors by leveraging multiple grid levels $\Omega_{2^{l-1}h}$, $l=1,\dots,L$. Here, $\Omega_{2^{l-1}h}$ is a uniform grid on $[0,1]$ with mesh size $2^{l-1}h$, and we assume $1/h$ is an integer divisible by $2^{L-1}$. Part of the Low-frequency components in the error on the fine mesh $\Omega_{2^{l-1}h}$ will become high-frequency after being projected to the coarser mesh $\Omega_{2^lh}$, and therefore easier for relaxation methods to solve. At the coarsest level ($l=L$), the system is small enough to be solved exactly via direct methods (e.g., Gaussian elimination). By alternating between relaxation methods and coarse-grid correction, MG ensures that all error frequencies are addressed.

\begin{figure}
    \centering
    \includegraphics[width=1.\linewidth]{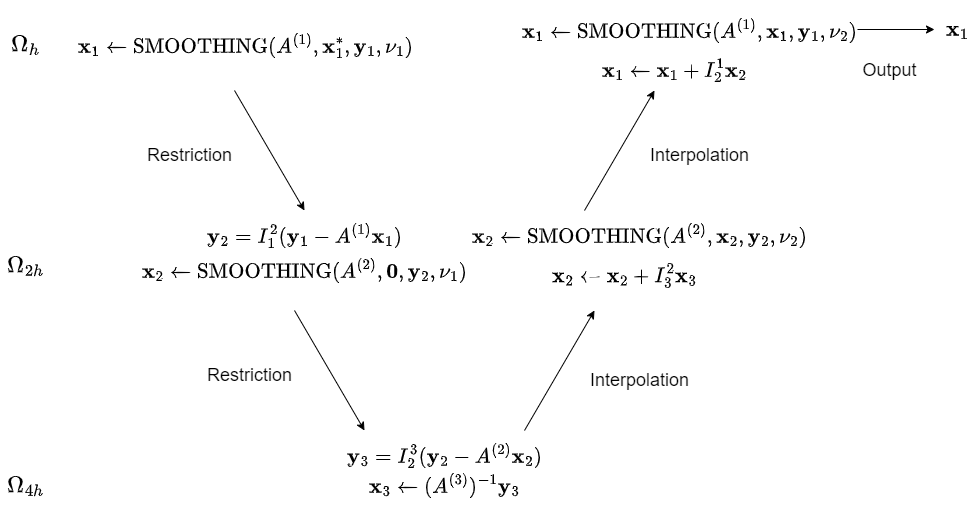}
    \caption{Graph illustration of a V-cycle multigrid scheme with $L=3$}
    \label{fig:MG_graph}
\end{figure}

A complete multigrid cycle (V-cycle) for solving $A\bx=\by$ with an initial guess $\bx^*$ is defined in Algorithm \ref{alg:multigrid} and a graph illustration is given in Figure \ref{fig:MG_graph}. We denote it as \[\text{MG}(A,\bx^*,\by,\nu_1,\nu_2,1,L).\] The MG function contains two main components, namely, the smoothing step and the coarsest grid correction step. The smoothing steps are performed by a smoother on $\Omega_{2^{l-1}h}$ for $l<L$, dumping the high-frequency in the error, while the coarsest grid correction step is performed by an exact solver at $\Omega_{2^{L-1}h}$, eliminating the geometric low-frequency components.
At $l=1,\dots, L-1$, we first perform $\nu_1$ iterations of pre-smoothing steps $\text{SMOOTHING}(A^{(l)},\bx_l^*,\by_l,\nu_1)$, that is, solving $A^{(l)}\bx_l=\by_l$ via classical relaxation methods with an initial guess $\bx_l^*$. Then, we restrict the residual $\by_l - A^{(l)} \bx_l$ to the next grid $\Omega_{2^lh}$ by multiplying a restriction matrix $I_{l}^{l+1}$ and get $\br_{l+1}$, which is further used as the right-hand-side vector of the coarser grid system. At $l=L$, we solve the coarsest grid linear system $A^{(L)}\bx_L=\by_L$ by an exact solver and return the solution.
After finishing all computations in coarser grids, we interpolate the coarser grid solution $\be_{l+1}=\text{MG}(A^{(l+1)},\mathbf 0, \br_{l+1}, \nu_1, \nu_2, l+1, L)$ to the grid level $l$ by multiplying an interpolation matrix $I_{l+1}^{l}$, and apply $\nu_2$ iterations of post-smoothing steps $\text{SMOOTHING}(A^{(l)},\bx_l, \by_l, \nu_2)$. More details about the algorithm and analysis of multigrid methods can be found in \cite{xu2017algebraic}. Finally, the multigrid method solves $A\bx=\by$ by iteratively applying multigrid cycles to refine the solution:
\[ \bx^{\tau+1}=\text{MG}(A,\bx^{\tau},\by,\nu_1,\nu_2,1,L),\ \tau=0,1,\dots,\ \bx^0=\mathbf 0.\]

\begin{algorithm}[h]
\caption{Multigrid cycle (V-cycle)}
\label{alg:multigrid}
\begin{algorithmic}[1]
\Procedure{MG}{$A^{(l)}$, $\bx_l^*$, $\by_l$, $\nu_1$, $\nu_2$, $l$, $L$}
    \State \textbf{Input:} Coefficient matrix $A^{(l)}$, initial guess $\bx_l$, right-hand side $\by_l$, number of pre-smoothing steps $\nu_1$, number of post-smoothing steps $\nu_2$, current level $l$, and the coarsest level $L$
    \State \textbf{Output:} Improved solution $\bx_l$
    
    \If{on the coarsest grid $l=L$}
        \State $\bx_L = (A^{(L)})^{-1}\by_L$  \Comment{Coarse grid solver}
        \State \textbf{return} $\bx_L$
    \Else
        \State $\bx_l \gets \text{SMOOTHING}(A^{(l)},\bx_l^*, \by_l, \nu_1)$ \Comment{Pre-smoothing}
        
        \State $\by_{l+1} \gets I_{l}^{l+1} (\by_l - A^{(l)} \bx_l)$ 

        \State $A^{(l+1)} \gets I_{l}^{l+1} A^{(l)} I^{l}_{l+1}$ 

        \State $\bx_l \gets \bx_l + I_{l+1}^l\text{MG}(A^{(l+1)},\mathbf 0, \by_{l+1}, \nu_1, \nu_2, l+1, L)$ \Comment{$\mathbf 0$ is an all-zero vector}
        
        \State $\bx_l \gets \text{SMOOTHING}(A^{(l)},\bx_l, \by_l, \nu_2)$ \Comment{Post-smoothing}
        
        \State \textbf{return} $\bx_l$
    \EndIf
\EndProcedure
\end{algorithmic}
\end{algorithm}

To look at an example, consider solving a one-dimensional PDE problem described in Section \ref{sec:smoothers}. At the first grid $\Omega_h$, as we have shown in Section \ref{sec:smoothers}, the smoother can efficiently reduce Fourier modes $\be_\phi$ for $\phi\in\Phi_1:=[-\pi,-\pi/2)\cup(\pi/2,\pi]$, since the reduction factor $1-\mu_\phi$ is at least $1/2$ for $\phi\in\Phi_1$. Then, the restriction operator projects the unresolved low-frequency mode $\phi\in[-\pi/2,\pi/2]$ to the second grid $\Omega_{2h}$. Half of the low-frequency in the first grid, i.e., $\Phi_2:=[-\pi/2,-\pi/4)\cup(\pi/4,\pi/2]$, becomes high-frequency at the second grid level, so $\Phi_2$ would be efficiently damped by smoothers at this grid level. The remaining unresolved part $[-\pi/4,\pi/4]$ is further projected to a coarser grid. Generally, we define
\begin{equation}
    \Phi_l:=\begin{cases}
        [-\pi/2^{l-1},-\pi/2^{l})\cup(\pi/2^{l},\pi/2^{l-1}], & l=1,\dots,L-1\\
        [-\pi/2^{l-1},\pi/2^{l-1}], & l=L
    \end{cases}\label{eq:phi_definition1d}.
\end{equation}
At the $l$-th grid level, the smoother can efficiently reduce frequency components $\phi\in\Phi_l$ in the error, and the unresolved part $\cup_{k=l+1}^L\Phi_k$ is left to coarser grids. 
Finally,  $\Phi_L$ is projected to the coarsest grid and solved by an exact solver. For two-dimensional PDEs, similar analysis can be found in \cite{briggs2000multigrid}, and the two-dimensional frequency domain $[-\pi,\pi]\times[-\pi,\pi]$ can also be decomposed similarly:
\begin{equation}
    \hat\Phi_l:=\begin{cases}
        \{(\phi_1,\phi_2)|\max\{|\phi_1|,|\phi_2|\}\in(\pi/2^l,\pi/2^{l-1}]\}, & l=1,\dots,L-1\\
        [-\pi/2^{l-1},\pi/2^{l-1}]\times [-\pi/2^{l-1},\pi/2^{l-1}], & l=L
    \end{cases}\label{eq:phi_definition2d}.
\end{equation}
\begin{figure}[h]
    \centering
    \subfigure[One-dimensional problem]{\includegraphics[width=0.4\textwidth]{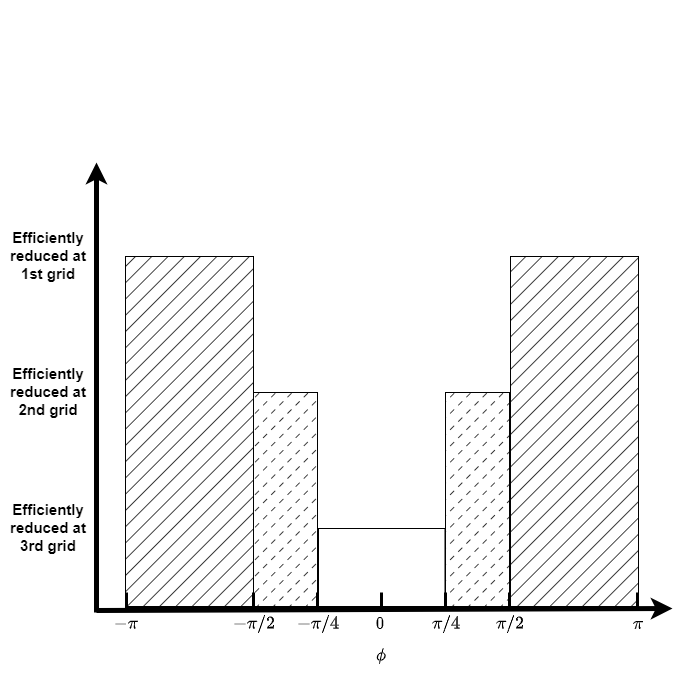}}
    \subfigure[Two-dimensional problem]{\includegraphics[width=0.4\textwidth]{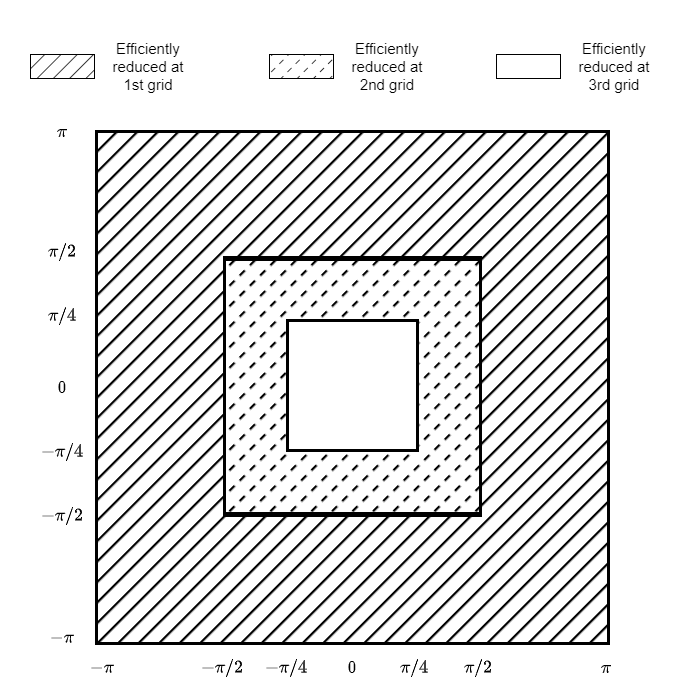}}
    \caption{Explanation of the frequency domain decomposition (L=3).}
    \label{fig:frequency_expalin}
\end{figure}
We further visualize the decomposition of frequency domains in Figure \ref{fig:frequency_expalin} for both one- and two-dimensional problems when $L=3$.

We remark that the multigrid method would fail for integral equations. It is because, unlike most PDE problems, the smoothing steps solve the same low-frequency part as the coarsest grid correction steps (see Figure \ref{fig:jacobi_iterations}), and leave the high-frequency parts unresolved. Detailed experiments and analysis of solving integral equations using multigrid methods can be found in \cite{chan1997multigrid}.

\section{The neural multigrid method}

% To solve convolution-type integral equations within the multigrid framework, 
To solve convolution-type integral equations efficiently, it is essential to identify an effective smoother capable of eliminating high-frequency error components. Inspired by recent advances incorporating neural operators into multigrid methods \cite{huang2022learning,Zhang2024BlendingNO,zhang2022hybrid}, we propose leveraging neural operators for smoothing steps in Algorithm \ref{alg:multigrid} to solve convolution-type integral equations. We refer to these neural operators employed for smoothing as neural smoothers. This section first briefly introduces neural operators, then presents our modified multigrid algorithms for both one-dimensional and two-dimensional convolution-type integral equations, and finally discusses the training strategies for the neural smoothers.

\subsection{Neural operators} \label{sec:neural_operators}
Neural operators are a specialized class of neural networks designed to approximate solution operators for mathematical problems, particularly partial differential equations (PDEs) \cite{lu2021learning,kovachki2023neural,li2024new}. Recently, there has been growing interest in solving linear systems using neural operators \cite{kahana2023geometry, Zhang2024BlendingNO,huang2022learning,chen2025graph,li2025solving,hu2025hybrid,kopanivcakova2025deeponet,chen2022meta}. Broadly, neural operators can be categorized into two types: The first type represents the solution in the continuous form \cite{lu2021learning,wang2021learning}, while the other type represents the solution on a discretized grid \cite{li2021fourier,kovachki2023neural}. Given that solutions of linear systems are inherently discrete vectors, we focus primarily on the latter type.
Various conventional neural network architectures are applicable to this task, including fully connected networks, ResNet \cite{he2016deep}, U-Net \cite{ronneberger2015u}, GCN \cite{thomas2016semi}, and the Fourier neural operator (FNO) \cite{li2021fourier}. In this work, we adopt the FNO architecture for implementing the methods proposed in subsequent sections because of its success in a wide range of applications \cite{li2021fourier,wen2022u,li2023fourier}.

The FNO is defined as the composition of multiple Fourier layers:
\begin{align*}
    \mathcal{N}_\theta(\by):=\tilde{f}_L\circ\tilde{f}_{L-1}\dots\circ \tilde{f}_1(\by)
\end{align*}
where
\begin{align*}
    \tilde{f}_l(\bz):=\sigma_l(W_l\bz+\text{DFT}^{-1}\circ R_l\circ \text{DFT}(\bz)), l=1,\dots,L.
\end{align*}
Here, $\sigma_l$ is the nonlinear activation function, $\text{DFT}$ is the Fourier transformation, $\text{DFT}^{-1}$ is the inverse Fourier transformation, and $W_l$ and $R_l$ are learnable linear operators. Specifically, $R_l$ is usually parameterized as a complex-valued weight matrix, and $W_l$ can be parameterized as a convolution operator. 
% We also visualize the structure of FNO in Figure \ref{fig:FNO}.

% \begin{figure}[]
%     \centering
%     \includegraphics[width=0.8\textwidth]{figures/FNO strucutres.drawio.png}
%     \caption{Illustration of the structure of FNO.}
%     \label{fig:FNO}
% \end{figure}

Typically, the training of neural operators can be data-driven or physics-informed. The data-driven method directly minimizes the difference between the network prediction $\mathcal{N}_\theta(\by)$ and the true solution $\bx=A^{-1}\by$:
\begin{equation*}
    \min_\theta \frac{1}{N}\sum_{i=1}^N\|\mathcal{N}_\theta(\by_i)-\bx_i\|_2^2.
\end{equation*}
Here $\{\bx_i,\by_i=A\bx_i\}_{i=1}^N$ is a set of training samples. The physics-informed method \cite{wang2021learning,li2024physics,goswami2022physics} minimizes the residual of the problem:
\begin{equation*}
    \min_\theta \frac{1}{N}\sum_{i=1}^N\|A\mathcal{N}_\theta(\by_i)-\by_i\|_2^2.
\end{equation*}
The physics-informed method is particularly useful when the true solution $\mathbf x_i=A^{-1}\mathbf y_i$ is unavailable. 

\subsection{Neural multigrid method for one-dimensional integral equation}
We first consider the linear system 
\[A\bx=\by,\]
where $A$ is the coefficient matrix of a one-dimensional integral equation (\ref{eq:integral equation}).
We will replace the smoothing steps in Algorithm \ref{alg:multigrid} at each grid $\Omega_{2^{l-1}h}$ with a neural smoother $N_{\theta_l}:\mathbb R^{n_l}\rightarrow\mathbb R^{n_l}$, $l=1,\dots,L-1$. Here, $n_l$, $l=1,\dots,L-1$, denotes the length of the residual vector at the $l$-th grid. For simplicity, we only perform the presmoothing steps, \textit{i.e.}, set $\nu_2=0$ (This reduces the V-cycle to the backslash ("$\backslash$") cycle \cite[Algorithm 5.14]{xu1996introduction}). The proposed multigrid algorithm is described in Algorithm \ref{alg:multigrid_NO}. Except for the smoothing steps, the other parts of Algorithm \ref{alg:multigrid_NO} are identical to Algorithm \ref{alg:multigrid}.

\begin{algorithm}[ht]
\caption{Neural multigrid algorithm for one-dimensional integral equation}
\label{alg:multigrid_NO}
\begin{algorithmic}[1]
\Procedure{\text{NMG1d}}{$\bx_l^*$, $\by_l$, $A^{(l)}$, $l$, $L$}
    \State \textbf{Input:} Coefficient matrix $A^{(l)}$, initial guess $\bx_l^*$, right-hand side $\by_l$, current level $l$
    \State \textbf{Output:} Improved solution $\bx_l$
    
    \If{on the coarsest grid $l=L$}
        \State $\bx_L = (A^{(L)})^{-1}\by_L$  \Comment{Coarse grid solver}
        \State \textbf{return} $\bx_L$
    \Else
        \State $\mathbf b_l\gets \by_l-A^{(l)}\bx_l^*$
        \State $\mathbf h_l^{\theta_l}\gets N_{\theta_l}(\mathbf b_l/\|\mathbf b_l\|_2)\|\mathbf b_l\|_2$ \Comment{Pre-smoothing with neural smoothers}
        \State $\bx_l\leftarrow \bx_l^*+\mathbf h_l^{\theta_l}$
         \State $\by_{l+1} \gets I_{l}^{l+1} (\by_l - A^{(l)} \bx_l)$ 

        \State $\bx_l \gets \bx_l + I_{l+1}^{l}\text{NMG1d}(\mathbf 0, \by_{l+1}, I_{l}^{l+1} A^{(l)} I^{l}_{l+1}, l+1, L)$
        
        \State \textbf{return} $\bx_{l}$
    \EndIf
\EndProcedure

\end{algorithmic}
\end{algorithm}

To train all neural smoother $\{N_{\theta_l}\}_{l=1}^{L-1}$, we need to define appropriate loss functions. Suppose we have a set of randomly generated training data $\{(\bx_i, \by_i)\}_{i=1}^N$ where $\by_i=A\bx_i$. A natural choice is using the squared error of the entire multigrid cycle:
\begin{equation}
    \mathcal{L}(\theta_1,\dots,\theta_{L-1})=\frac{1}{N}\sum_{i=1}^N \left\| \bx_i-\text{NMG1d}(\mathbf 0,\by_i,A,1,L)\right\|^2_2. \label{eq:loss_1}
\end{equation}
The loss function (\ref{eq:loss_1}) is also used in learning the multigrid smoother for PDE problems \cite{huang2022learning}.
While this loss function offers a simple and unified training objective across all neural smoothers, its joint optimization introduces a lack of control over the learning dynamics of individual neural smoothers. Specifically, the aggregated squared error fails to isolate the contributions of each smoother, potentially resulting in imbalanced learning dynamics. For instance, some neural smoothers may converge rapidly while others converge slowly during training. Additionally, since the entire multigrid cycle is optimized together, changing the coarsest level $L$ at the testing stage will substantially impact the performance of the neural multigrid method. These shortcomings suggest that more refined level-wise training losses are necessary to ensure stability and interpretability for neural smoothers.

In this work, we aim to design neural smoothers capable of efficiently solving the high-frequency components at each grid level and leaving the low-frequency components to coarser levels. This behavior is similar to the conventional multigrid smoother when solving PDE problems. To achieve this, we constrain neural smoothers $N_{\theta_l}$ to focus on reducing high-frequency errors corresponding to $\Omega_{2^{l-1}h}$, {\it i.e.}, $\phi\in\Phi_l$, where $\Phi_l$ is defined as (\ref{eq:phi_definition1d}). 
Following the notation of Algorithm \ref{alg:multigrid_NO}, we denote the output vector of $N_{\theta_l}$ as $(\mathbf h_l^{\theta_l})_i$ and the right-hand-side vector at the coarsest grid level as $(\by_L)_i$ for the $i$-th training sample $(\bx_i,\by_i)$. Then, one cycle of the neural multigrid algorithm gives
\[ \text{NMG1d}(\mathbf 0,\by_i,A,1,L)=\sum_{l=1}^{L-1}I_{l}^{1}(\mathbf h_l^{\theta_l})_i+I_L^1(A^{(L)})^{-1}(\by_L)_i.\]
In the neural multigrid algorithm, neural smoothers are applied sequentially. At the $l$-th grid level, the approximation given by the previous $l-1$ neural smoothers is $\sum_{k=1}^{l-1}I_k^1(\mathbf h_k^{\theta_k})_i$ and the remaining unresolved error is $\mathbf{x}_i - \sum_{k=1}^{l-1}I_k^1(\mathbf h_k^{\theta_k})_i$. Based on previous analysis, $N_{\theta_l}$ is expected to efficiently reduce the frequency components in $\Phi_l$ and leave the remaining frequency components $\cup_{k=l+1}^L\Phi_k$ to coarser grids. Consequently, we consider training $N_{\theta_l}$ with the following loss function:
\begin{align}
    L_l(\theta_l):=\frac{1}{N}\sum_{i=1}^N \left\|\mathbb F_l\left( \bx_i-\sum_{k=1}^{l-1}I_{k}^{1}(\mathbf h_k^{\theta_k})_i-I_{l}^{1}(\mathbf h_l^{\theta_l})_i\right)\right\|^2_2
    \label{eq:loss_2},
\end{align}  
where $\mathbb F_l$ is a frequency filter acting on the frequency domain that assigns large weights to high-frequency components $\Phi_l$ and small or zero weights to low-frequency components:
\[\mathbb F_l(\cdot):=\mathbf m_l\odot\text{DFT}_{1}(\cdot).\]
Here $\text{DFT}_{1}$ is the one-dimensional discrete Fourier transform, $\mathbf m_l\in\mathbb R^{n_1}$ is a mask acting on the frequency domain, and $\odot$ denotes the Hadamard product. The choice of $\mathbf m_l$ would significantly affect the convergence of different frequency components in the training error. To see this, we can analyze different frequency components in the training error.

Let $(\bx_l')_i:=\bx_i-\sum_{k=1}^{l-1}I_{k}^{1}(\mathbf h_k^{\theta_k})$ be the remaining unresolved error at the $l$-th grid for the $i$-th sample. Then, $(\bx_l')_i-I_l^1(\mathbf h_l^{\theta_l})_i$ represent the prediction error of the neural smoother $N_{\theta_l}$ for the $i$-th sample. We define $f_i^j:=|\text{DFT}_1((\bx_l')_i-I_l^1(\mathbf h_l^{\theta_l})_i)[j]|^2$ as the squared magnitude of the $j$-th frequency component of the error vector $(\bx_l')_i-I_l^1(\mathbf h_l^{\theta_l})_i$, and $\mathcal F_j:=\sum_{i=1}^Nf_i^j$ as the sum of 
all $f_i^j$ over the entire training set. We further denote $F_l\in\mathbb C^{n_l\times n_l}$ as the $n_l$-dimensional normalized DFT matrix and $D_l=\text{diag}(\mathbf m_l)\in\mathbb R^{n_l\times n_l}$ be a diagonal matrix whose diagonal vector is $\mathbf m_l$. Then, the loss (\ref{eq:loss_2}) can be rewritten as 
\[L_l=\frac{1}{N}\sum_{i=1}^N\Vert D_lF_l((\bx_l')_i-I_l^1(\mathbf h_l^{\theta_l})_i)\Vert_2^2=\frac{1}{N}\sum_{i=1}^N\sum_{j=1}^{n_l}\mathbf m_l[j]^2f_i^j=\sum_{j=1}^{n_l}\mathbf m_l[j]^2\mathcal F_j.\]
During training, when $\mathbf m_l[j]> 0$, we can derive an upper bound of $\mathcal F_j$ as
\[\mathcal F_j\leq \frac{L_l(\theta_l)}{\mathbf m_l[j]^2}.\]
We see that a larger $\mathbf m_l[j]$ would lead to a smaller upper-bound for $\mathcal F_j$ during training. Besides, this upper bound does not depend on the spectral property of the coefficient matrix $A$. Therefore, by properly defining $\mathbf m_l$ for $l=1, \dots, L-1$, we can train $N_{\theta_l}$ to solve the corresponding high-frequency components for all kinds of linear systems, including both PDE and integral equations.

\begin{figure}[]
    \centering
    \subfigure[$m_1$]{\includegraphics[width=0.3\textwidth]{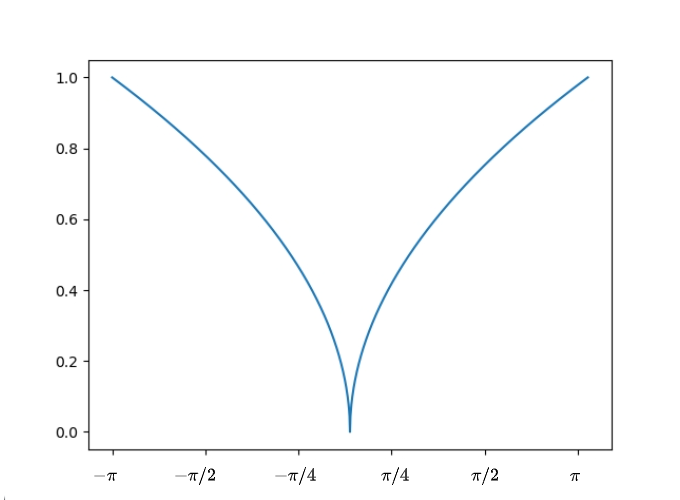}}
    \subfigure[$m_2$]{\includegraphics[width=0.3\textwidth]{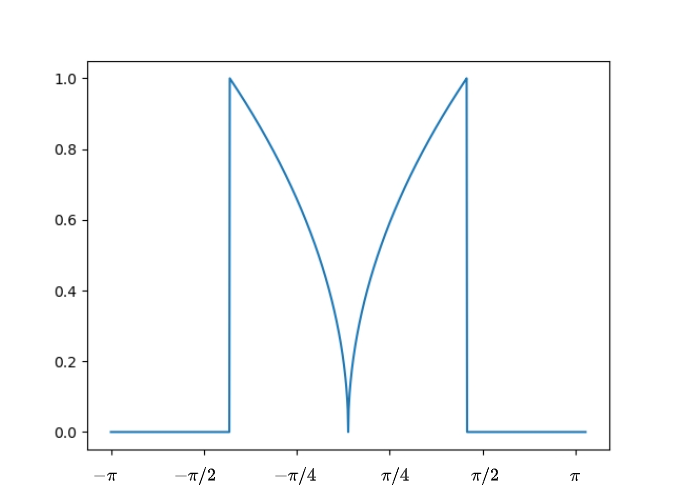}}
    \subfigure[$m_3$]{\includegraphics[width=0.3\textwidth]{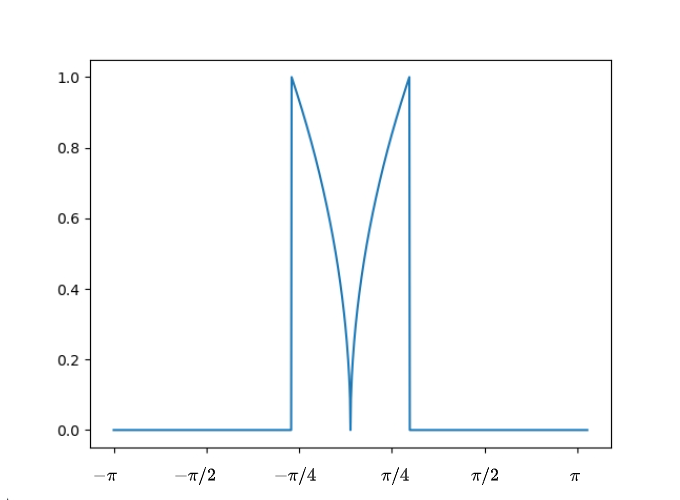}}
    \caption{Plot of $\mathbf m_l$ for $l=1,2,3$.}
    \label{fig:1d_mask}
\end{figure}

In our previous analysis of the weighted Jacobi method in Section \ref{sec:smoothers}, we see that the reduction factor for frequency $\phi\in[-\pi,\pi]$ is $1-\mu_\phi$ (see Figure \ref{fig:smoothing factor}). The factor $1-\mu_\phi$ achieves the maximum at the highest frequency $\phi=\pm\pi$ and gradually decreases to 0 when $\phi\rightarrow0$. We would choose $\mathbf m_l$ following the same idea. We first define a continuous function $m_l:[-\pi,\pi]\rightarrow\bbR$ such that 
\[m_l(\phi):=\begin{cases}
    \sqrt{\frac{\phi}{\pi/2^{l-1}}}, & \phi\in[-\pi/2^{l-1},\pi/2^{l-1}]\\
    0, & \text{else}
\end{cases}.\]
$m_l(\phi)$ obtain the maximum at $|\phi|=\pi/2^{l-1}$ which is the highest frequency visible at the $l$-th grid, and gradually decrease to 0 when $\phi\rightarrow0$. Here, $m_l(\phi)=0$ when $\phi$ is not visible at the $l$-th grid. Then, $\mathbf m_l$ is obtained by discretizing $m_l$ at a uniform mesh over $[-\pi,\pi]$ with $n_l$ grid points.
We also visualize $m_l$ for $l=1,2,3$ in Figure \ref{fig:1d_mask}. 

\begin{figure}
    \centering
    \includegraphics[width=.9\linewidth]{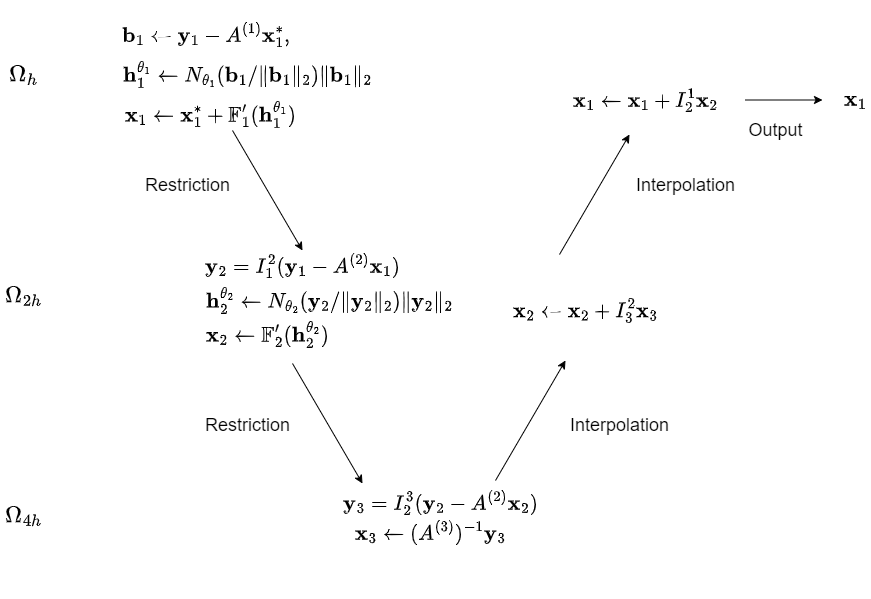}
    \caption{The graph illustration of the neural multigrid cycle for one-dimensional problems when $L=3$.}
    \label{fig:nmg1d}
\end{figure}

Since $N_{\theta_l}$ focuses on reducing frequency components in $\Phi_l$, its prediction error may be large at the other frequency components. Therefore, to improve the stability of the neural multigrid cycle, we consider adding similar frequency filters to Algorithm \ref{alg:multigrid_NO} as well. 
At the $l$-th grid, we introduce another frequency filter $$\mathbb F_l'(\cdot):=\text{Re}(\text{DFT}^{-1}_{1}(\mathbf{m}_l'\odot\text{DFT}_{1}(\cdot))),$$ 
where $\mathbf{m}_l'\in\mathbb R^{n_l}$ is the uniform discretization of $m_l$ over $\cup_{k=l}^L\Phi_k=[-\pi/2^{l-1},\pi/2^{l-1}]$ which is the frequency range visible at the $l$-th grid, and $\text{Re}(\cdot)$ is the real part function. The main difference between $\mathbb F_l$ and $\mathbb F_l'$ is that $\mathbb F_l$ operates on the first (finest) grid while $\mathbb F_l'$ operates on the $l$-th grid. We then apply $\hat{\mathbb F}_l$ to $h_l^{\theta_l}$ in Algorithm \ref{alg:multigrid_NO}, which gives a modified neural multigrid cycle (Algorithm \ref{alg:multigrid_NO_modified}). We also present a graph illustration of this neural multigrid cycle in Figure \ref{fig:nmg1d} when $L=3$. Applying $\mathbb F'_l$ to $h_l^{\theta_l}$ mainly keeps its high-frequency components and eliminates its low-frequency parts.
The training loss (\ref{eq:loss_2}) shall also be modified as
\begin{equation}
    \mathcal{L}_l(\theta_l;\{\theta_k\}_{k=1}^{l-1}):=\frac{1}{N}\sum_{i=1}^N \left\|\mathbb F_l\left( \bx_i-\sum_{k=1}^{l-1}I_{k}^{1}\mathbb F_k'((\mathbf h_k^{\theta_k})_i)-I_{l}^{1}(\mathbf h_l^{\theta_l})_i\right)\right\|^2_2
    \label{eq:loss_2_modified}.
\end{equation}
After training, the neural multigrid algorithm solves the one-dimensional integral equations by iteratively applying the neural multigrid cycle:
 \[\bx^{\tau}=\text{NMGF1d}(A,\bx^{\tau-1},\by,1,L),\ \tau=1,2,\dots,\ \bx^0=\mathbf 0.\]

\begin{algorithm}[ht]
\caption{Neural multigrid algorithm with frequency filter for one-dimensional integral equations}
\label{alg:multigrid_NO_modified}
\begin{algorithmic}[1]
\Procedure{\text{NMGF1d}}{$\bx_l^*$, $\by_l$, $A^{(l)}$, $l$, $L$}
    \State \textbf{Input:} Coefficient matrix $A^{(l)}$, initial guess $\bx_l^*$, right-hand side $\by_l$, current level $l$
    \State \textbf{Output:} Improved solution $\bx_l$
    
    \If{on the coarsest grid $l=L$}
        \State $\bx_L = (A^{(L)})^{-1}\by_L$ 
        \State \textbf{return} $\bx_L$
    \Else
        \State $\mathbf b_l\gets \by_l-A^{(l)}\bx_l^*$
        \State $\mathbf h_l^{\theta_l}\gets N_{\theta_l}(\mathbf b_l/\|\mathbf b_l\|_2)\|\mathbf b_l\|_2$ \Comment{Pre-smoothing with neural smoothers and frequency filter}
        \State $\bx_l\leftarrow \bx_l^*+\mathbb F_l'(\mathbf h_l^{\theta_l})$
        
         \State $\by_{l+1} \gets I_{l}^{l+1} (\by_l - A^{(l)} \bx_l)$ \

        \State $\bx_l \gets \bx_l + I_{l+1}^{l} \text{NMGF1d}(\mathbf 0, \mathbf y_{l+1},I_{l}^{l+1} A^{(l)} I^{l}_{l+1}, l+1, L)$
        
        \State \textbf{return} $\bx_{l}$
    \EndIf
\EndProcedure
\end{algorithmic}
\end{algorithm}

\subsection{Neural multigrid method for two-dimensional integral equations}

Now we consider a two-dimensional integral equation
\begin{equation}
    \alpha \hat R(u)(z_1,z_2)+\int_{\Omega^2}  \hat{\mathcal{K}}(z_1-z_1',z_2-z_2')u(z_1',z_2')dz_1'dz_2'=\hat f(z_1,z_2),\quad (z_1,z_2)\in[0,1]^2. \label{eq:integral equation2d}
\end{equation}
where $\hat R$ is a two-dimensional regularization term, $\hat{\mathcal K}$ is a two-dimensional convolution kernel, and $\hat f$ is a right-hand-side function. Discretizing two-dimensional functions $u$ and $f$ will yield two matrices $X$ and $Y$. We further define $\text{Vec}$ as the vectorization transformation that transforms a matrix into a vector by concatenating its column vectors sequentially. For example, let $X$ be an $n\times m$ matrix, then
    \[\text{Vec}(X)=(X_{11},\dots,X_{n1},X_{12},\dots,X_{n2},\dots,X_{1m},\dots,X_{nm})^\top,\]
where $X_{ij}$ denotes the element at $i$-th row and $j$-th column of $X$.
The inverse of $\text{Vec}$ is the matricization function which is denoted as $\text{Mat}:=\text{Vec}^{-1}$. Then, the discretized system of (\ref{eq:integral equation2d}) would be given as
\begin{equation}
    \hat A\text{Vec}(X)=\text{Vec}(Y) \label{eq:system_2d},
\end{equation}
where $\hat A$ is the corresponding coefficient matrix.
Equation (\ref{eq:system_2d}) can be directly solved using a neural multigrid method like Algorithm \ref{alg:multigrid_NO_modified}, but extra modifications are needed. 

\begin{figure}[ht]
    \centering
    \subfigure[$\hat{M}_1$]{\includegraphics[width=0.3\textwidth]{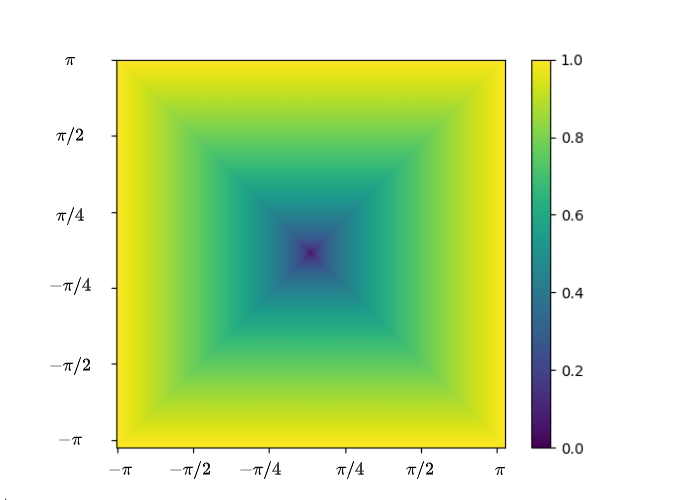}}
    \subfigure[$\hat{M}_2$]{\includegraphics[width=0.3\textwidth]{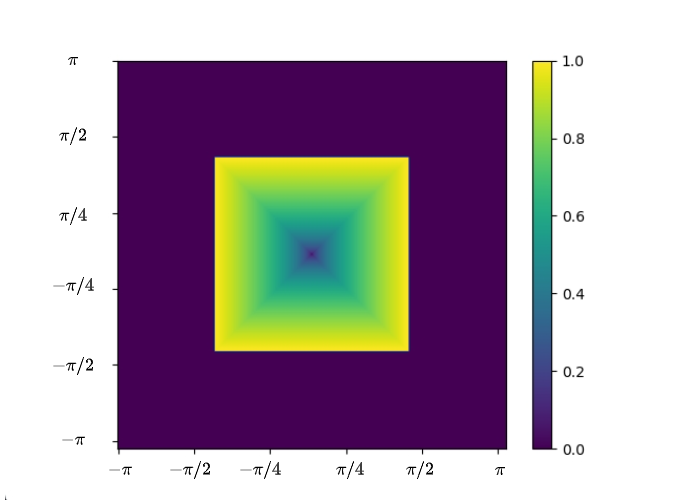}}
    \subfigure[$\hat{M}_3$]{\includegraphics[width=0.3\textwidth]{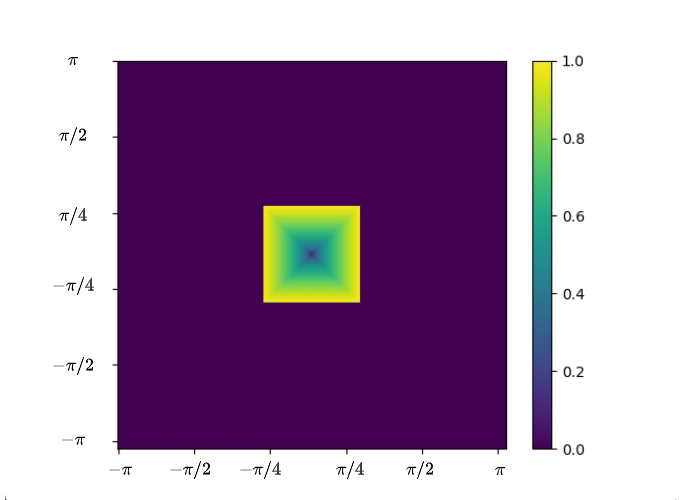}}
    \caption{Plot of $\hat{M}_l$ for $l=1,2,3$.}
    \label{fig:2d_mask}
\end{figure}

We present the two-dimensional version of Algorithm \ref{alg:multigrid_NO_modified} in Algorithm \ref{alg:multigrid_NO_2d}. The overall structure of Algorithm \ref{alg:multigrid_NO_2d} is the same as Algorithm \ref{alg:multigrid_NO_modified}. For two-dimensional problems, we define neural smoothers on the matrix spaces $\hat N_{\theta_l}:\bbR^{n_l\times m_l}\rightarrow \bbR^{n_l\times m_l}$, where $n_l\times m_l$ is the size of the residual matrix $R_l$ at the $l$-th grid. 
\begin{algorithm}[]
\caption{Neural multigrid algorithm with frequency filters for two-dimensional integral equations}
\label{alg:multigrid_NO_2d}
\begin{algorithmic}[1]
\Procedure{NMGF2d}{$X_l^*$, $Y_l$, $\hat A^{(l)}$, $l$, $L$}
    \State \textbf{Input:} Coefficient matrix $A^{(l)}$, initial guess $X_l^*$, right-hand-side $Y_l$, current level $l$
    \State \textbf{Output:} Improved solution $X_l$
    
    \If{on the coarsest grid $l=L$}
        \State $X_L = \text{Mat}\left(\left(\hat A^{(L)}\right)^{-1}\text{Vec}(Y_L)\right)$  
        \State \textbf{return} $X_L$
    \Else
        \State $B_l\gets Y_l-\text{Mat}\left(\hat A^{(l)}\text{Vec}(X_l^*)\right)$
        \State $H_l^{\theta_l}\gets \hat N_{\theta_l}(B_l/\|B_l\|_2)\|B_l\|_2$ 
        \State $X_l\gets X_l^*+\hat{\mathbb F}_l'(H_l^{\theta_l})$
        \State $R_{l} \gets Y_l - \text{Mat}\left(\hat A^{(l)}\text{Vec}(X_l)\right)$ 

        \State $R_{l+1} \gets \text{Mat}\left(\hat{I}_{l}^{l+1}\text{Vec}(R_l)\right)$ 

        \State $E_{l+1} \gets \text{NMGF2d}(O, R_{l+1}, \hat{I}_{l}^{l+1} \hat A^{(l)} \hat{I}^{l}_{l+1}, l+1, L)$ \Comment{$O$ is an all-zero matrix of the same size as $R_{l+1}$}
        
        \State $X_l \gets X_l + \text{Mat}\left(\hat{I}_{l+1}^{l}\text{Vec}(E_{l+1})\right)$
        
        \State \textbf{return} $X_{l}$
    \EndIf
\EndProcedure

\end{algorithmic}
\end{algorithm}
Similar to the one-dimensional cases, we define the frequency masks for two-dimensional problems. 
We first define a continuous function on a two-dimensional frequency domain as
\begin{equation*}
    \hat{m}_l(\phi_1,\phi_2)=\begin{cases}
    \sqrt{\frac{\max\{|\phi_1|,|\phi_2|\}}{\pi/2^{l-1}}}, & (\phi_1,\phi_2)\in [-\pi/2^{l-1},\pi/2^{l-1}]\times[-\pi/2^{l-1},\pi/2^{l-1}]\\
    0, & \text{else}
\end{cases}.
\end{equation*}
Here, $\hat{m}_l$ achieve the maximum value when $\max\{|\phi_1|,|\phi_2|\}=\pi/2^{l-1}$, which is the highest frequency visible at the $l$-th grid, and gradually decrease to 0 when $\max\{|\phi_1|,|\phi_2|\}\rightarrow 0$.
Then, we define frequency filters as
\[ \hat{\mathbb{F}}_l(\cdot)=\hat{M}_l\odot \text{DFT}_2(\cdot), \]
where $\text{DFT}_2$ is the two-dimensional discrete Fourier transform, and $\hat{M}_l\in\bbR^{n_1\times m_1}$ is a uniform discretization of $\hat m_l$ over $[-\pi,\pi]\times[-\pi,\pi]$,
and 
\[\hat{\mathbb F}_l'(\cdot):=\text{Re}(\text{DFT}_2^{-1}(\hat{M}_l'\odot\text{DFT}_2(\cdot))),\]
where $\hat{M}_l'\in\bbR^{n_l\times m_l}$ is the uniform discretization of $\hat m_l$ over $\cup_{k=l}^L\hat\Phi_l=[-\pi/2^{l-1},\pi/2^{l-1}]\times[-\pi/2^{l-1},\pi/2^{l-1}]$. The filter $\hat{\mathbb{F}}_l$ is applied at the finest grid, while $\hat{\mathbb{F}}_l$ is applied at the $l$-th grid.
Consequently, the loss function for $\hat N_{\theta_l}$ is defined as
\begin{align}
    &\hat{\mathcal{L}}_l(\theta_l;\{\theta_k\}_{k=1}^{l-1})\notag\\
    =&\frac{1}{N}\sum_{i=1}^N \left\|\hat{\mathbb F}_l\left( X_i-\sum_{k=1}^{l-1}\text{Mat}\left(\hat{I}_{k}^{1}\text{Vec}(\hat{\mathbb F}_k'(H_k^{\theta_k}))\right)-\text{Mat}\left(\hat{I}_{l}^{1}\text{Vec}(H_l^{\theta_l})\right)\right)\right\|^2_2 \label{eq:loss_3}
\end{align}   
which is the two-dimensional counterpart of (\ref{eq:loss_2_modified}).
We also visualize $M_l$ in Figure \ref{fig:2d_mask} for $l=1,2,3$.

\subsection{Training strategy}
After properly defining loss functions (\ref{eq:loss_2_modified}), we present the general training algorithm for one-dimensional integral equations in Algorithm \ref{alg:training}. Following standard neural network training procedures, we randomly generate small batches of training data. For each batch, we evaluate loss functions and compute parameter gradients through backpropagation. We then update network parameters using the Adam optimizer \cite{kingma2014adam}. To enhance the long-term stability of our neural multigrid algorithm, we randomly perform several cycles of neural multigrid during training data generation. The resultant residual and error serve as training samples. These neural multigrid evaluations utilize the latest neural smoothers without participating in backpropagation computations. The two-dimensional version of Algorithm \ref{alg:training} follows a similar approach and is therefore omitted. All training is conducted on a device with NVIDIA RTX A6000 (48GB) GPU and an Intel Core i9-10980XE (3.00 GHz) CPU.

\begin{algorithm}[]
\caption{Training of neural smoothers}
\label{alg:training}
\begin{algorithmic}[1]
\Procedure{Train}{$A$, $L$, $K$, $N_{bs}$, $N_{epoch}$}
    \State \textbf{Input:} The coefficient matrix $A$, maximum level $L$, maximum number of rollout iteration $K$, batch size $N_{bs}$, and number of epochs $N_{epoch}$
    \State \textbf{Output:} Neural smoothers $\{N_{\theta_l}\}_{l=1}^{L-1}$
    \State Initialize neural smoothers $\{N_{\theta_l}\}_{l=1}^{L-1}$
    \For{$\be$ from $1$ to $N_{epoch}$}
    \State $\{\bx_i,\by_i\}_{i=1}^{N_{bs}}\gets \text{DataGeneration}(N_{bs},A,L,K)$
    \State Evaluate $\text{NMGF1D}(\mathbf 0,\by_i,A,1,L)$
    \State Compute $\mathcal{L}_l(\theta_l;\{\theta_k\}_{k=1}^{l-1})$ for $l=1,\dots,L-1$ 
    \State $\theta_l\gets\theta_l-\eta_l\nabla_{\theta_l}\mathcal{L}_l(\theta_l;\{\theta_k\}_{k=1}^{l-1})$,\quad $l=1,\dots,L$
    \EndFor
\EndProcedure
\Procedure{DataGeneration}{$N_{bs}$, $A$, $L$, $K$}
    \State \textbf{Output:} $\{\bx_i,\by_i\}_{i=1}^{N_{bs}}$
    \For{$i$ from $1$ to $N_{bs}$}
    \State Generate $\bx_i$ from a given distribution randomly and compute $\by_i=A\bx_i$
    \State Generate random integer $k$ from $1$ to $K$
    \For{$j$ from $1$ to $k$}
    \State $\bx_i\gets \bx_i-\text{NMGF1d}(\mathbf 0, \by_i; A, 1, L)$
    \State $\by_i\gets \by_i-A\bx_i$
    \EndFor
    \EndFor
\EndProcedure
\end{algorithmic}
\end{algorithm}

\section{Numerical experiments}

\subsection{One-dimensional problems} \label{sec:1d_integral_exp}
\subsubsection{Integral equation with the Tikhonov regularization}
We first consider equation (\ref{eq:integral equation}) where $R(u)=u$, $\mathcal{K}$ is the Gaussian smoothing kernel, and $\Omega=[0,1]$. The $R(u)=u$ corresponds to the gradient of the Tikhonov regularization term $\| u\|_{2}^2$. We discretize (\ref{eq:integral equation}) on a uniform mesh $\Omega_h$ with mesh size $h$, and denote $n=1/h$.
The coefficient matrix of the discretized system is
\begin{equation}
    A=\alpha I+K \label{eq:1d_linear_system}
\end{equation}
where $I$ is the identity matrix, and $K$ is a convolution matrix. We set the standard deviation of the Gaussian kernel to be 1.5. 
For the training of neural smoothers (Algorithm \ref{alg:training}), we choose $L=4$, $N_{bs}=20$, $K=10$, and $N_{epoch}=1500$. 
Then, we train our neural smoothers with different choices of $\alpha$ and $n$, and compare them with the backslash multigrid scheme (Algorithm \ref{alg:multigrid} with $L=4$, $\nu_1=5$, and $\nu_2=0$). Details of the structure of neural smoothers are presented in \ref{sec:network_structure}. To accelerate the convergence of training, we use the idea of curriculum learning \cite{bengio2009curriculum}. The coefficient matrix $A$ with a smaller $\alpha$ would have a larger condition number, and it would be more difficult to train neural smoothers. Therefore, we can first train neural smoothers on an easy problem ($A$ with large $\alpha$), then we use these neural smoothers to initialize the training of difficult problems ($A$ with small $\alpha$), which can significantly accelerate the convergence.
% We show the condition number of $A$ and the required training time for neural smoothers in Table \ref{tab:condNum1d}. 

During the evaluation, we test on 10 randomly generated right-hand-side vectors. We show the averaged number of cycles, the standard deviation of the number of cycles, and the averaged evaluation time required to meet the stopping criterion (when the relative residual reaches $10^{-6}$) in Table \ref{tab:1dIntegral}. We see that our neural multigrid algorithm converges much faster than the classical multigrid algorithm. Besides, the iteration number required by our method is less sensitive to the changes in $\alpha$ and $n$.

\begin{table}[]
\centering
\begin{tabular}{|>{\centering}p{0.23\textwidth}|>{\centering}p{0.1\textwidth}|>{\centering}p{0.17\textwidth}|>{\centering}p{0.17\textwidth}|>{\centering}p{0.17\textwidth}|}
\hline
   &        &  $\alpha=10^{-4}$ & $\alpha=10^{-5}$ & $\alpha=10^{-6}$ \tabularnewline \hline
\multirow{3}{*}{\makecell{proposed method}}       & $n=256$ & 13.7(0.17s)$\pm 1.8$ & 13.7(0.18s)$\pm 2.5$   & 14.6(0.19s)$\pm 0.8$\tabularnewline   
                                                 & $n=512$ & 13.3(0.21s)$\pm 0.6$ & 13.1(0.20s)$\pm 0.9$ & 14.2(0.24s)$\pm 0.6$ \tabularnewline   
                                                 & $n=1024$ & 14.0(0.28s)$\pm 1.2$ & 13.4(0.27s)$\pm 1.7$ & 14.4(0.30s)$\pm 1.0$ \tabularnewline
                                                 %   
                                                 % & $6-\text{level}$ & 14.4(0.50s)$\pm 1.56$ & 14.2(0.49s)$\pm 0.75$ & 13.4(0.46s)$\pm 0.92$ \tabularnewline
                                                 \hline
% \multirow{3}{*}{\makecell{iteration number of the\\multigrid (V-cycle) method}}       & $\alpha=10^{-3}$ & 3460(2.4s) & 3463(3.9s) & 3471(7.8s) \tabularnewline   
%                                                   & $\alpha=10^{-4}$ & 26694(16.2s) & 26733(30.5s) & 26787(60.1s)\tabularnewline   
%                                                   & $\alpha=10^{-5}$ & 84956(57.2s) & 85147(97.7s) & 85505(193.3s)
%  \tabularnewline \hline
 \multirow{3}{*}{\makecell{multigrid method}}       & $n=256$ & 9556(7s)$\pm 139$ & 27423(16s)$\pm 542$   & 35375(21s)$\pm 699$\tabularnewline   
                                                 & $n=512$ & 9570(16s)$\pm 87$ & 27484(45s)$\pm 330$ & 35455(58s)$\pm 425$ \tabularnewline   
                                                 & $n=1024$ & 9589(56s)$\pm 73$ & 27600(162s)$\pm 286$ & 35604(210s)$\pm 369$ \tabularnewline
                                                 %   
                                                 % & $6-\text{level}$ & 5122(10.2s)$\pm 90$ & 11334(22.6s)$\pm 286$ & 13178.6(26.1s)$\pm 364$ \tabularnewline
                                                 \hline
\end{tabular}
\caption{The average number of iterations and evaluation time required for solving the one-dimensional integral equation.}
\label{tab:1dIntegral}
\end{table}

% \begin{table}[]
% \centering
% \begin{tabular}{@{}ccccc@{}}
% \toprule
%                            & 6-level & 5-level & 4-level & 3-level \\ \midrule
% $\alpha=10^{-6}$ & 13.4(0.92)    & 12.9(0.94)    & 12.6(0.80)    & 11(0.89)      \\
% $\alpha=10^{-5}$ & 14.2(0.75)    & 13(0.63)      & 11.7(0.64)    & 12(1.3)      \\
% $\alpha=10^{-4}$ & 14.4(1.56)    & 13.5(1.36)    & 12.2(1.07)    & 11(0.63)      \\ \bottomrule
% \end{tabular}
% \caption{Convergence iterations of neural multigrid algorithm}
% \label{tab:1d_results}
% \end{table}

\begin{figure}
    \centering
    \includegraphics[width=.8\linewidth]{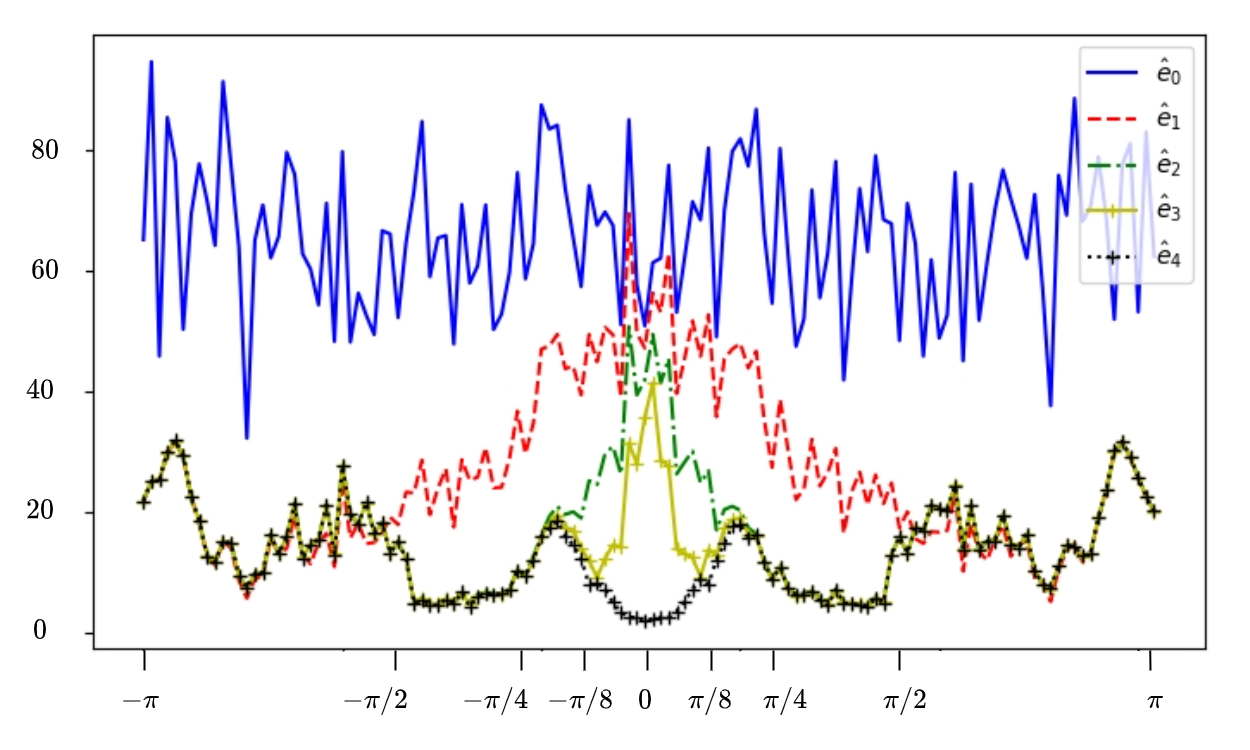}
    \caption{Magnitude spectrums of unsolved errors at the fine grid after applying neural smoothers at each level.}
    \label{fig:1d_magnitude_spectrum}
\end{figure}

For any given true solution $\bx$ and the corresponding right-hand-side vector $\by=A\bx$, let $\mathbf h_l^{\theta_l}$ be the output of the neural smoother $N_{\theta_l}$ when evaluating $\text{NMGF1D}(\mathbf{0},\hat{\by};A,1,L)$. We define $\be_l$ as the remaining error after the smoothing step at the $l$-th grid level:
\[ \be_l:=\bx-\sum_{k=1}^l\mathbf I_{k}^1\mathbb F_k'(\mathbf h_k^{\theta_k}), \] 
and define $\hat{\be}_l$ as the magnitude spectrum of $\be_l$. The change from $\hat{\be}_{l-1}$ to $\hat{\be}_l$ can reflect which part of the frequency components is solved by $N_{\theta_l}$.  In Figure \ref{fig:1d_magnitude_spectrum}, we visualize the magnitude spectra $\hat{\be}_l$ for different $l$ in one neural multigrid cycle ($L=4$), where $\hat{\be}_0$ corresponds to the magnitude spectrum of the true solution and $\hat{\be}_4$ corresponds to the magnitude spectrum of the error after the coarse grid correction. From Figure \ref{fig:1d_magnitude_spectrum}, we see that the original $\hat{\be}_0$ contains frequency components across the entire spectrum.
The frequency $\phi\in\Phi_1$ is significantly reduced in $\hat{\be}_1$ after applying $N_{\theta_1}$. Then, $\Phi_2$ and $\Phi_3$ are further reduced in $\hat{\be}_2$ and $\hat{\be}_3$ respectively after applying $N_{\theta_2}$ and $N_{\theta_3}$. Finally, the lowest frequency part $\Phi_4$ is efficiently resolved in $\be_4$ by the coarse grid correction. It demonstrates that our neural smoothers perform like the classical relaxation methods for PDE problems.

\subsubsection{Integral equation with the anisotropic regularization}
Here, we also consider another regularization term 
$R(u)(z)=-\nabla\cdot(a(z)u(z))$
for the integral equation (\ref{eq:integral equation}), where $a(z)=1+0.5\sin(2\pi z)$. The coefficient matrix of the discretized system is $A=\alpha D+K$, where $D$ corresponds to the regularization term and $K$ corresponds to the convolution term. Here, $K$ remains consistent with the previous subsection, and $D$ is a tri-diagonal matrix with varying coefficients along the diagonals. We maintain identical training and testing configurations as in the Tikhonov case. We compare our methods with the multi-grid method for this problem with different $n$ and $\alpha$, and the results are presented in Table \ref{tab:1dIntegral_2}. Similar to the results in Table \ref{tab:1dIntegral}, our neural multigrid algorithm converges faster than the multigrid algorithm, and the iteration number of our method is more robust to the changes in $\alpha$ and $n$.

\begin{table}[]
\centering
\begin{tabular}{|>{\centering}p{0.23\textwidth}|>{\centering}p{0.1\textwidth}|>{\centering}p{0.17\textwidth}|>{\centering}p{0.17\textwidth}|>{\centering}p{0.17\textwidth}|}
\hline
   &        &  $\alpha=10^{-4}$ & $\alpha=10^{-5}$ & $\alpha=10^{-6}$ \tabularnewline \hline
\multirow{3}{*}{\makecell{proposed method}}       & $n=256$ & 13.2(0.15s)$\pm 1.2$ & 13.8(0.16s)$\pm 0.9$ & 14.4(0.19s)$\pm 1.3$\tabularnewline   
                                                 & $n=512$ & 13.4(0.21s)$\pm 1.3$ & 14.0(0.23s)$\pm 1.1$ & 14.5(0.25s)$\pm 0.8$ \tabularnewline   
                                                 & $n=1024$ & 13.4(0.27s)$\pm 1.1$ & 13.4(0.26s)$\pm 1.0$ & 14.7(0.32s)$\pm1.3 $ \tabularnewline
                                                 \hline

 \multirow{3}{*}{\makecell{multigrid method}}    &  $n=256$  & 5130(4s)$\pm$173 & 20325(13s)$\pm$1034 & 31707(21s)$\pm$1613\tabularnewline   
                                                 & $n=512$ & 5184(9s)$\pm$121 & 20603(36s)$\pm$728 & 32140(56s)$\pm$1136 \tabularnewline   
                                                 & $n=1024$ & 5207(30s)$\pm$70 & 20742(118s)$\pm$420 & 32355(184s)$\pm$655\tabularnewline
                                                 \hline
\end{tabular}
\caption{The average number of iterations and evaluation time required for solving the one-dimensional integral equation.}
\label{tab:1dIntegral_2}
\end{table}

\subsubsection{Second-order PDE}
Although our method is primarily designed for integral equations, its framework applies to more general linear systems. To verify the generalizability of our approach, we test it on a simple one-dimensional PDE:  
\begin{equation}  
10^{-4}u(z) + u''(z) = f(z), \quad z \in [0,1], \label{eq:pde1d}  
\end{equation}  
subject to homogeneous Dirichlet boundary conditions \( u(0) = u(1) = 0 \).  

The training and testing configurations remain identical to those in earlier experiments, with numerical results summarized in Table \ref{tab:1dpde}.  
The multigrid method is more efficient than our proposed method for this PDE problem.
However, the main purpose of this experiment is to show the ability of our method to solve more general linear systems.
Here, our proposed method remains effective, and the required number of cycles is similar to the integral equations cases.

\begin{table}[]
\centering
\begin{tabular}{|>{\centering}p{0.23\textwidth}|>{\centering}p{0.17\textwidth}|>{\centering}p{0.17\textwidth}|>{\centering}p{0.17\textwidth}|}
\hline
   &   $n=256$ & $n=512$ & $n=1024$ \tabularnewline \hline
proposed method       & 14.0(0.17s)$\pm 0.6$ & 13.0(0.21s)$\pm 1.8$ & 13.7(0.28s)$\pm 1.8$\tabularnewline\hline

multigrid method   & 8(0.00s)$\pm 0$ & 8(0.01s)$\pm 0$ & 8(0.04s)$\pm 0$\tabularnewline
                                                 \hline
\end{tabular}
\caption{The average number of iterations (cycles) and evaluation time required for solving the one-dimensional PDE}
\label{tab:1dpde}
\end{table}

\subsection{Two-dimensional integral equation}
Here, we also consider the two-dimensional integral equation (\ref{eq:integral equation2d}). We first discretize the domain $[0,1]^2$ using a uniform grid $\Omega_h\times\Omega_h$ into $n^2$ cells where $n=1/h$. Then, the discretization of (\ref{eq:integral equation2d}) over the grid $\Omega_h\times\Omega_h$ yields a linear system whose coefficient matrix $\hat A$ admits the Kronecker product structure:
$$\hat A=\alpha I\otimes I+K\otimes K,$$
where $\otimes$ denotes the Kronecker product, and $I$ and $K$ are the identity matrix and the convolution matrix defined in (\ref{eq:1d_linear_system}). We compare our proposed method with the multigrid method and the conjugate gradient (CG) method across different $\alpha$ and $n$.  
Numerical experiments are conducted with 10 randomly generated right-hand-side vectors, and the average iterations (cycles) required to achieve a relative residual less than $10^{-6}$ is reported in Table \ref{tab:2dIntegral}. As shown in the table, our neural multigrid method demonstrates superior performance to classical multigrid in both iteration count and computational time. While the proposed method requires more computational time than the conjugate gradient method for large values of $\alpha$, it shows greater robustness in iteration count when $\alpha$ decreases (corresponding to an increasing condition number of A). Notably, at $\alpha=10^{-5}$, our method significantly outperforms conjugate gradient in both iteration count and computational efficiency.

\begin{figure}
    \centering
    \subfigure[$\hat{E}_{0}-\hat{E}_1$]{\includegraphics[width=0.45\textwidth]{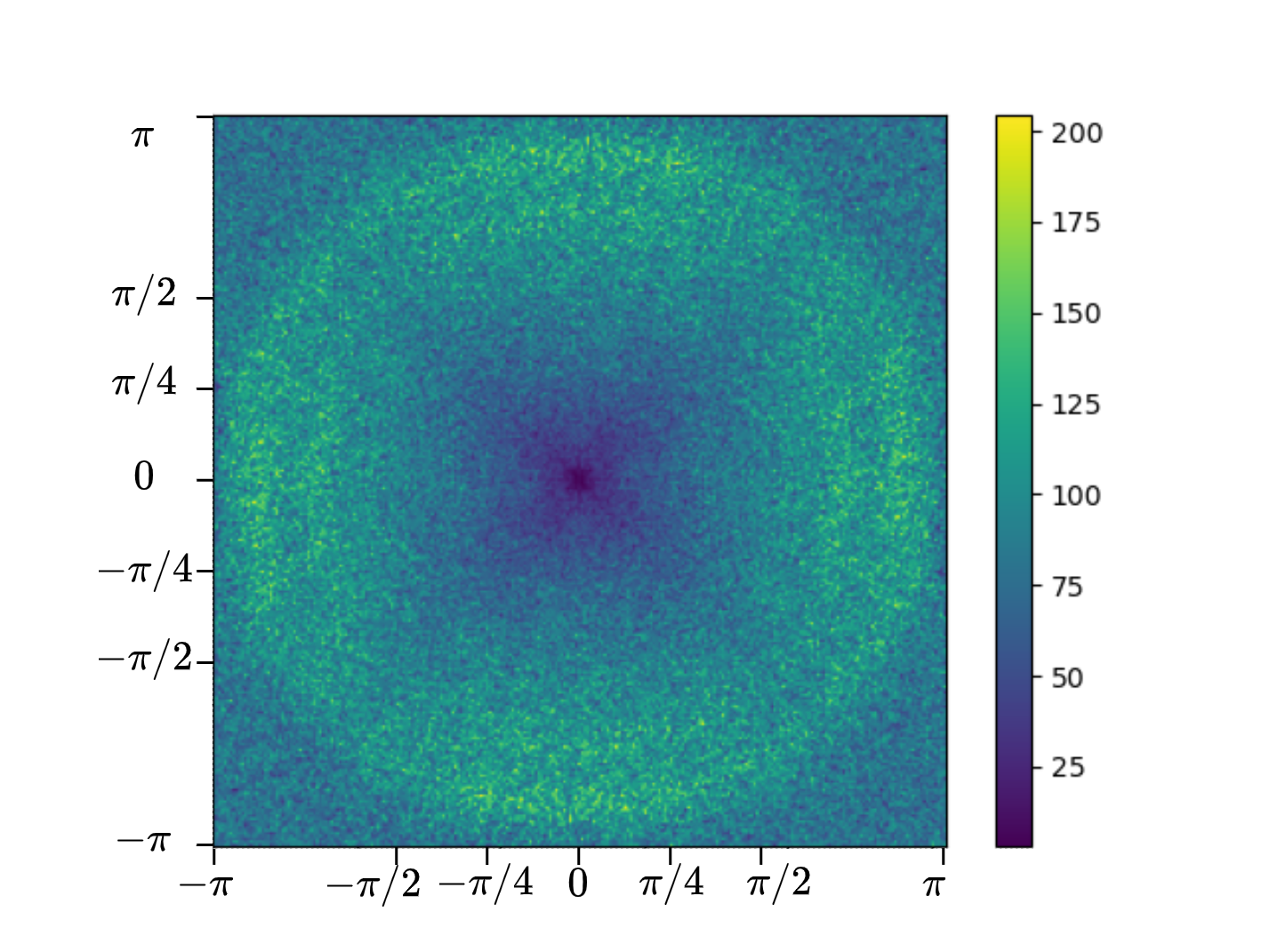}}
    \subfigure[$\hat{E}_{1}-\hat{E}_2$]{\includegraphics[width=0.45\textwidth]{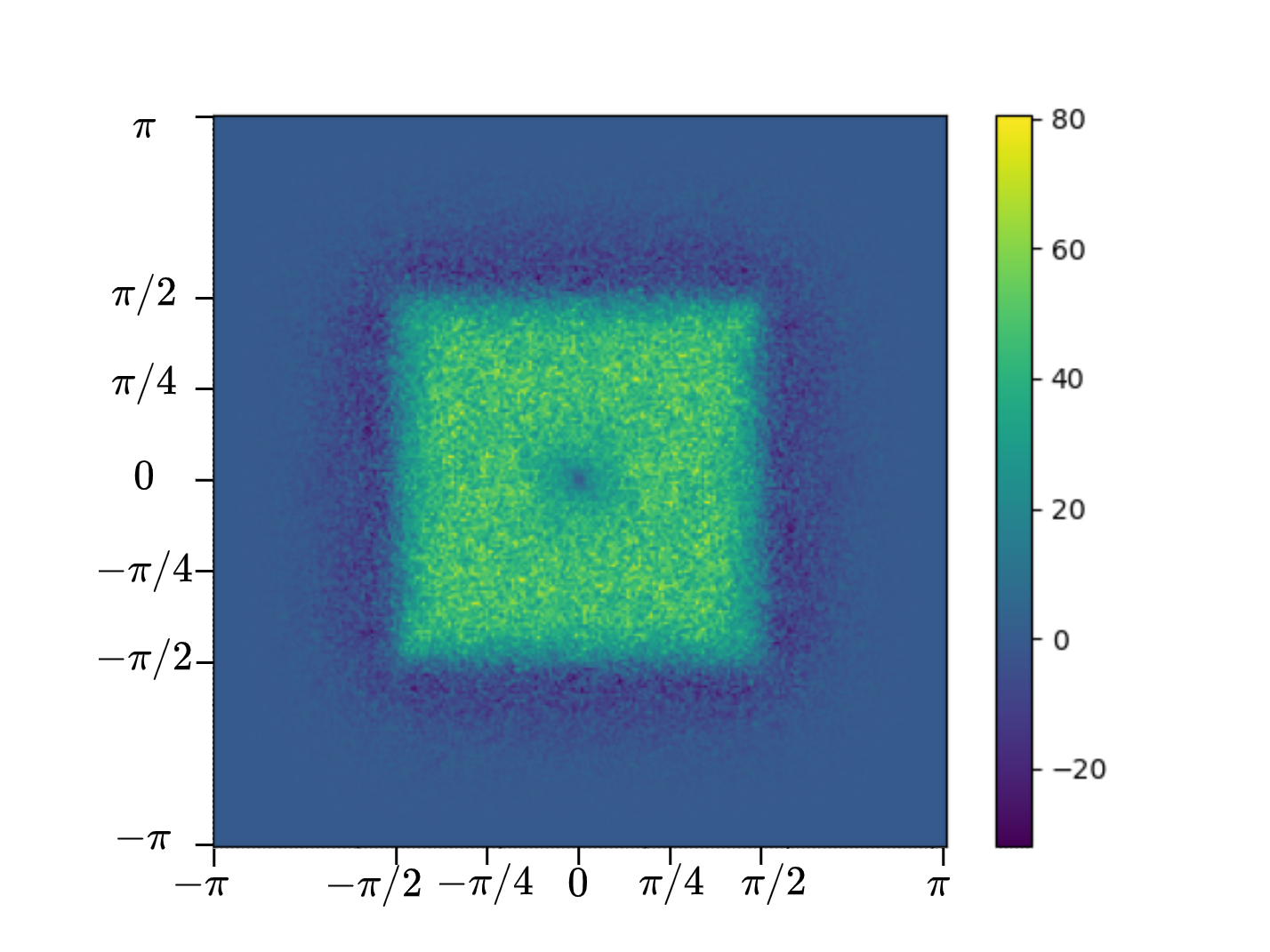}}
    \subfigure[$\hat{E}_{2}-\hat{E}_3$]{\includegraphics[width=0.45\textwidth]{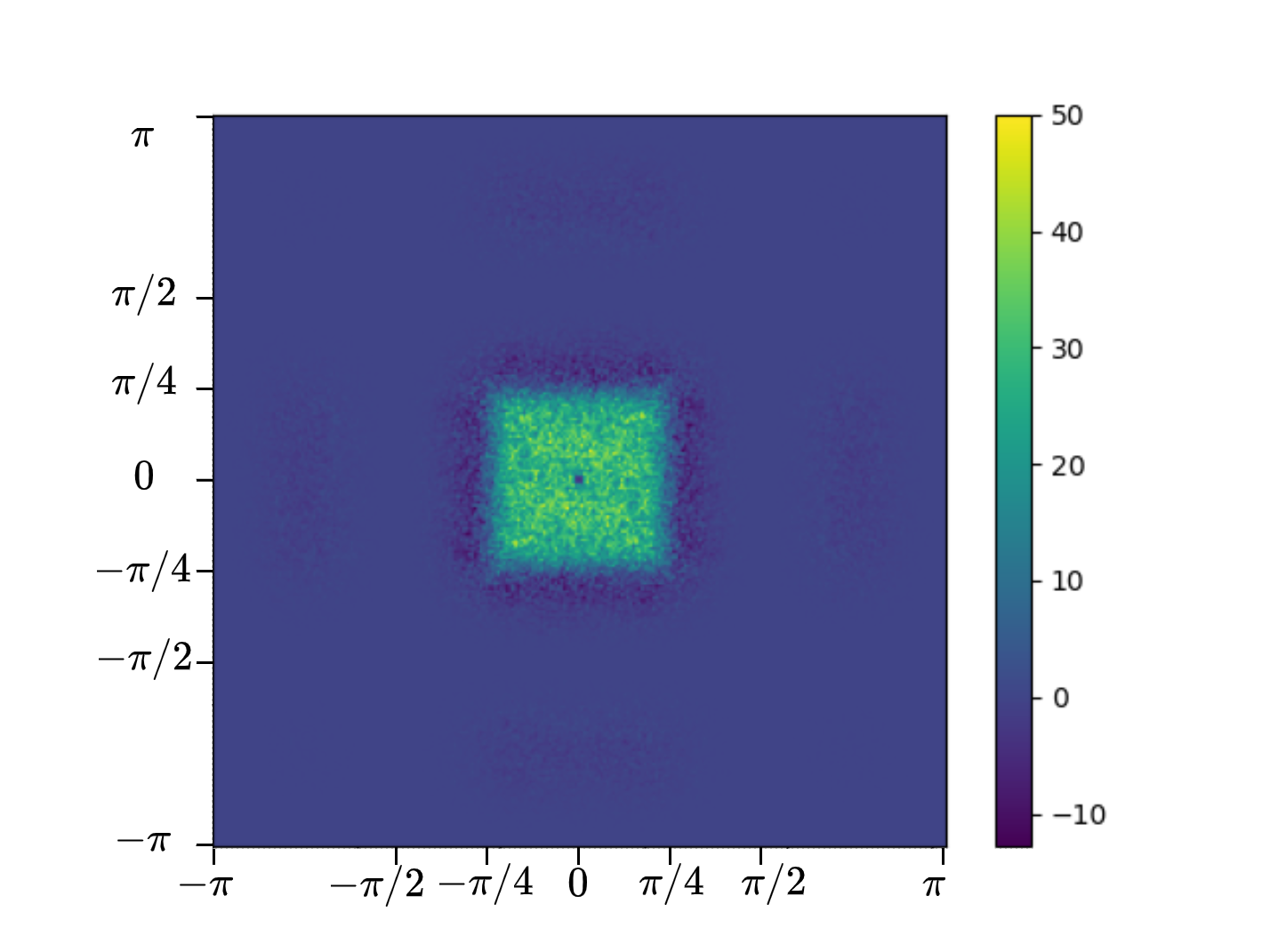}}
    \subfigure[$\hat{E}_{3}-\hat{E}_4$]{\includegraphics[width=0.45\textwidth]{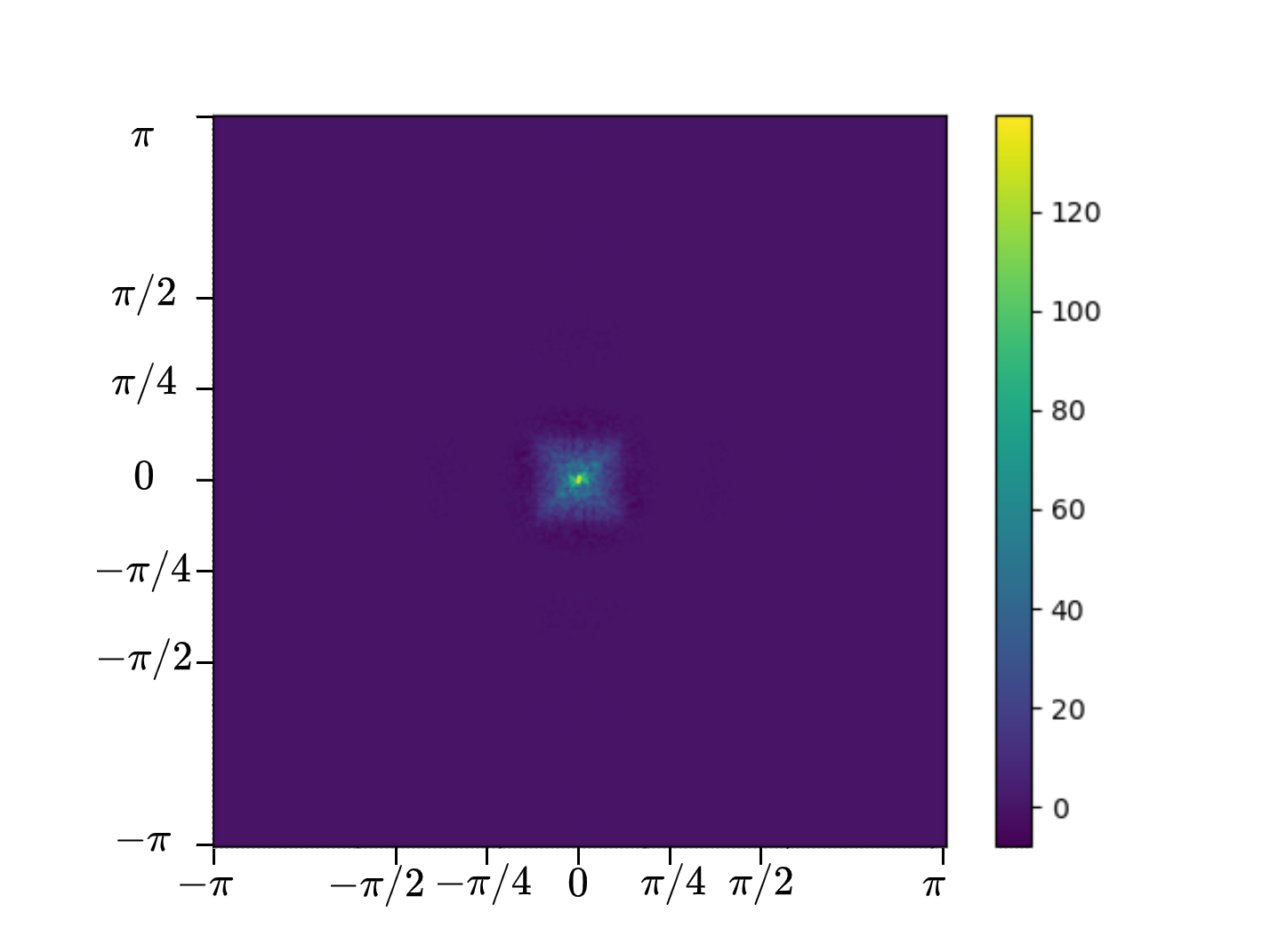}}
    \caption{Difference between $\hat{E}_l$ and $\hat{E}_{l-1}$ for $l=1,2,3,4$.}
    \label{fig:2D_spectrum_change}
\end{figure}

When solving a system $\hat A\text{Vec}(X)=\text{Vec}(Y)$ using the neural multigrid algorithm (Algorithm \ref{alg:multigrid_NO_2d}),
We define $E_l$ as the remaining error after applying the neural smoother at the $l$-th grid
\[E_l=X-\sum_{k=1}^{l}\text{Mat}\left(I_{k}^{1}\text{Vec}(\hat N_{\theta_k}(B_k))\right),\quad l=1,\dots,L-1,\]
$E_L$ as the remaining error after the coarse grid correction, and $E_0=X$ as the initial error.
We denote $\hat E_l$ as the magnitude spectrum of $E_l$. Then, the difference between $\hat{E}_l$ and $\hat{E}_{l-1}$ reveals which frequency components are effectively resolved by each neural smoother $\hat N_{\theta_l}$. We visualize the $\hat{E}_l-\hat{E}_{l-1}$ for $l=1,2,3,4$ in Figure \ref{fig:2D_spectrum_change}, which demonstrates that $\hat N_{\theta_l}$ successfully reduces the desired frequency components $\hat{\Phi}_l$ as expected.

\begin{table}[]
\centering
\begin{tabular}{|>{\centering}p{0.2\textwidth}|>{\centering}p{0.1\textwidth}|>{\centering}p{0.2\textwidth}|>{\centering}p{0.2\textwidth}|>{\centering}p{0.2\textwidth}|}
\hline
   &        &  $\alpha=10^{-3}$ & $\alpha=10^{-4}$ & $\alpha=10^{-5}$ \tabularnewline \hline
\multirow{3}{*}{\makecell{proposed method}}       & $n=256$ & $11.3$(4s)$\pm 2.2$ & $13.8$(5s)$\pm 3.0$   & $14.2$(5s)$\pm 3.7$\tabularnewline
                                                 & $n=512$ & $12.7$(12s)$\pm 4.5$ & $14.3$(13s)$\pm 2.4$ & $15$(14s)$\pm 4.5$ \tabularnewline  
                                                 & $n=1024$ & 13.6(55s)$\pm 4.3$ & 14.2(59s)$\pm 3.18$ & 14.6(61s)$\pm 2.2$ \tabularnewline
                                                 %   
                                                 % & $6-\text{level}$ & 15.2(s)$\pm 0.7$ & 15.3(s)$\pm 1.2$ & 15.2(s)$\pm 1.2$ \tabularnewline
                                                 \hline
% \multirow{3}{*}{\makecell{iteration number of the\\multigrid (V-cycle) method}}       & $\alpha=10^{-3}$ & 3460(2.4s) & 3463(3.9s) & 3471(7.8s) \tabularnewline   
%                                                   & $\alpha=10^{-4}$ & 26694(16.2s) & 26733(30.5s) & 26787(60.1s)\tabularnewline   
%                                                   & $\alpha=10^{-5}$ & 84956(57.2s) & 85147(97.7s) & 85505(193.3s)
%  \tabularnewline \hline
\multirow{3}{*}{\makecell{multigrid method}}       & $n=256$ & $1140$(17s)$\pm5$ & $9510$(2min)$\pm59$   & $71263$(18min)$\pm703$\tabularnewline
                                                 & $n=512$ &  $1145$(57s)$\pm4$ & $9555$(8min)$\pm 34$   & $77288$(1hr)$\pm397$ \tabularnewline 
                                                 & $n=1024$ &  $1146$(4min)$\pm2$ & $9577$(33min)$\pm18$ & $77310$(4hr)$\pm207$ \tabularnewline
                                                 %   
                                                 % & $6-\text{level}$ & 843(s)$\pm $ & 6995(s)$\pm $ & 57970(s)$\pm $ \tabularnewline
                                                 \hline
\multirow{3}{*}{\makecell{conjugate gradient}}       & $n=256$ & $178$(2s)$\pm2$ & $472$(5s)$\pm4$   & $1193$(12s)$\pm5$\tabularnewline 
                                                 & $n=512$ & $179$(7s)$\pm1$ & $476$(17s)$\pm2$   & $1208$(42s)$\pm2$ \tabularnewline 
                                                 & $n=1024$ & $179$(26s)$\pm1$ & $478$(64s)$\pm1$   & $1213$(158s)$\pm1$ \tabularnewline 
                                                 \hline
\end{tabular}
\caption{The average number of iterations and evaluation time required for solving the two-dimensional integral equation.}
\label{tab:2dIntegral}
\end{table}

\subsection{Comparison of different loss functions}
We compare the performance and stability of neural smoothers trained using the combined loss function (Eq. \ref{eq:loss_1}) and the proposed level-wise loss functions (Eq. \ref{eq:loss_2_modified}). The experiments are conducted on the previously introduced one-dimensional integral equation (\ref{eq:1d_linear_system}), with parameters \( L = 4 \), \( n = 1024 \), and \( \alpha = 10^{-4} \). Neural smoothers are trained separately using the two loss functions, while all other training and testing configurations remain consistent with those in Section \ref{sec:1d_integral_exp}.  

We also want to evaluate the robustness of the trained smoothers when changing the coarsest level during testing. We remain \( L = 4 \) during training, learning three distinct neural smoothers \( N_{\theta_1} \), \( N_{\theta_2} \), and \( N_{\theta_3} \). At testing time, we test varying coarsest grid levels \( L' = 2, 3, 4 \), \text{i.e.}, we only perform smoothing at the first \( L'-1 \) grids, then perform the coarsest grid correction at the \( L' \)-th level. The results are shown in Table \ref{tab:loss_comp}, where neural smoothers trained with the proposed level-wise losses (Eq. \ref{eq:loss_2_modified}) consistently require fewer cycles and exhibit a significantly lower standard deviation in all cases. Moreover, they demonstrate enhanced robustness to changes in \( L' \), aligning more closely with classical multigrid schemes.

\begin{table}[]
\centering
\begin{tabular}{|>{\centering}p{0.25\textwidth}|>{\centering}p{0.15\textwidth}|>{\centering}p{0.15\textwidth}|>{\centering}p{0.15\textwidth}|>{\centering}p{0.15\textwidth}|}
\hline
   &        &  $\alpha=10^{-4}$ & $\alpha=10^{-5}$ & $\alpha=10^{-6}$ \tabularnewline \hline
\multirow{3}{*}{\makecell{trained by the\\ level-wise losses (\ref{eq:loss_2_modified})}}       & $L'=2$ & $13.5\pm1.4$ & $13.4\pm2.0$   & $18.5\pm6.3$\tabularnewline   
                                                 & $L'=3$ & $12.5\pm0.8$ & $11.7\pm1.3$   & $12.6\pm1.7$ \tabularnewline   
                                                 & $L'=4$ & 14.0$\pm 1.2$ & 13.4$\pm 1.7$ & 14.4$\pm 1.0$ \tabularnewline
                                                 \hline
\multirow{3}{*}{\makecell{trained by the\\combined loss (\ref{eq:loss_1})}}       & $L'=2$ & $30.3\pm12.2$ & $32.6\pm9.7$   & $32.4\pm5.8$\tabularnewline   
                                                 & $L'=3$ & $15.4\pm6.7$ & $24.7\pm7.6$   & $24.2\pm5.1$ \tabularnewline   
                                                 & $L'=4$ & $16.8\pm5.5$ & $19.4\pm6.5$   & $16.6\pm6.3 $
                                                 \tabularnewline
                                                 \hline
\end{tabular}
\caption{The average number of cycles when training with different loss functions}
\label{tab:loss_comp}
\end{table}

\section{Conclusion}  
In this work, we propose a novel level-wise training strategy (LTS) for training neural operators to solve integral equations. Motivated by the behavior of the classical multigrid scheme solving PDE problems, we guide neural operators to focus on specific frequency components by incorporating specific frequency filters into the training loss functions. The resulting trained neural operators serve as smoothers in multigrid schemes, forming an efficient and robust neural multigrid method. Notably, our neural multigrid method demonstrates generalizability to other types of problems like PDEs as well.

A current limitation of our approach lies in its restriction to linear systems with fixed coefficient matrices, where only the right-hand-side vectors may vary. When coefficient matrices change, the method requires retraining of all neural smoothers. In future work, we plan to extend this framework to solve classes of linear systems with varying coefficient matrices. Another direction of the future work is exploring the possibility of using a single neural operator as smoothers at different levels, giving a more flexible multigrid scheme.

\bmhead{Acknowledgements.} This work is partially supported by InnoHK initiative of the Innovation and Technology Commission of the Hong Kong Special Administrative Region Government, HKRGC Grants No. CityU11309922, HKRGC Grants No. LU13300125, ITF Grant No. MHP/054/22, LU BGR 105824, and the Natural Sciences and Engineering Research Council of Canada

\bibliography{bibfile} 

\appendix
\renewcommand{\thesection}{Appendix \Alph{section}}
\section{Network structures} \label{sec:network_structure} 
In numerical experiments, we adopt Fourier neural operators (FNOs) \cite{li2021fourier} for both one- and two-dimensional problems. We first use a fully connected layer to lift the number of channels $c$. Next, we concatenate sequences of Fourier layers defined as  
$$  
\sigma\left(\text{DFT}^{-1} \circ R \circ \text{DFT}(Z) + WZ + B\right),  
$$  
where $Z$ is the input signal with $c$ channels to this layer, $\sigma$ is a nonlinear activation function, $R$ is a learnable complex linear transformation applied only to the first $k$ low-frequency components and damped other high-frequency components channelwisely, $W$ is a learnable linear transformation, and $B$ is a bias vector. The total number of Fourier layers is denoted as $L$. Finally, we use another fully connected layer to reduce the number of channels to one. Further implementation details of the FNO can be found in \cite{li2021fourier}.  
Here, $W$ is implemented as a convolution operator with learnable convolutional kernels. The hyperparameters of neural smoothers defined on different grids $\Omega_h$ (one-dimensional problem) or $\Omega_h \times \Omega_h$ (two-dimensional problems) are provided in Table \ref{tab:network_structures}.  

\begin{table}[]
\centering
\begin{tabular}{|c|c|cccc|}
\hline
problem dimension & $h$ & $c$ & $L$ & $k$ & kernel size\\ \hline
\multirow{5}{*}{one-dimensional}       & $1/64$        &  64   &  4   &  16   &     5                    \\      
                                        & $1/128$        &  64   &  4   &  32   &     5                    \\      
                                        & $1/256$        &  64   &  5   &  64   &     7                    \\
                                      & $1/512$        & 64    &  6  &  128   &  7                           \\
                                      & $1/1024$       & 64    &  6   & 128    &   7                         \\ \hline
\multirow{5}{*}{two-dimensional}        & $1/64$        &  64   &  4   &  16   &     3                    \\      
                                        & $1/128$        &  64   &  4   &  32   &     3                    \\          & $1/256$        &  64   &  5   &   32  &      5                   \\
                                      & $1/512$        &  64   &   6  & 64    &    7                         \\
                                      & $1/1024$       &  64   &  6   &  64   &    7                  \\\hline  
\end{tabular}
\caption{Hyperparameters of neural smoothers}
\label{tab:network_structures}
\end{table}

\end{document}